\documentclass[%
  aip,pof,
  amsmath,amssymb,%
  reprint,
  nofootinbib%
]{revtex4-1}

\usepackage{graphicx}
\usepackage{dcolumn}
\usepackage{bm}
\usepackage[utf8]{inputenc}
\usepackage[T1]{fontenc}
\usepackage{mathptmx}
\usepackage{etoolbox}

\usepackage{booktabs}
\usepackage{longtable}
\usepackage{array}

\usepackage{bibunits}
\defaultbibliographystyle{aipnum4-1}
\defaultbibliography{paper}
\usepackage{multirow}
\usepackage{siunitx}
\newcommand{\vect}[1]{\bm{#1}}
\newcommand{\GNS}{\textsc{gns}}
\newcommand{\ELGIN}{\textsc{elgin}}

\newcommand{\MzeroMDE}{19.56}        
\newcommand{\MzeroMDEatFive}{7.81}
\newcommand{\MzeroMDEatFifteen}{24.81}

\newcommand{\MzeroValidFrac}{87}
\newcommand{\MzeroLastValidTime}{25.0}
\newcommand{\MzeroLastValidMDE}{29.03}
\newcommand{\MzeroKE}{1.057}

\newcommand{\MzeroRg}{9.85}     

\newcommand{\elginMDE}{16.20}        
\newcommand{\elginMDEatFive}{7.19}
\newcommand{\elginMDEatFifteen}{17.75}
\newcommand{\elginMDEatTwentyEight}{22.26}
\newcommand{\elginKE}{0.659}

\newcommand{\elginRg}{6.58}      

\newcommand{\relMDEimprovement}{17}
\newcommand{\relRgimprovement}{33}

\newcommand{\peakBZECFD}{0.50}     
\newcommand{\peakBZEELGIN}{1.00} 
\newcommand{\peakBZEdiff}{0.50} 
\newcommand{\BZErmse}{0.25}             

\newcommand{\rolloutSec}{64}
\newcommand{\rolloutSecMzero}{27}
\newcommand{\rolloutTimeMzero}{\rolloutSecMzero}

\newcommand{\speedupX}{37}

\newcommand{\dpBarMicron}{21.7}

\newcommand{\StokesRange}{4.8\!\times\!10^{-5}\,\textendash\,2.4\!\times\!10^{-4}}

\begin{document}

\begin{bibunit}

\preprint{AIP/123-QED}

\title{Physics-Informed Graph Neural Network Surrogates for Turbulent
       Nanoparticle Dispersion in Dental Clinical Environments}

\author{Takshak Shende}
\email{takshak.shende@gmail.com}
\affiliation{Department of Mechanical Engineering, University College London (UCL),
             London, United Kingdom}
\altaffiliation{Current affiliation: Ascend Technologies Ltd,
                Southampton, United Kingdom}

\author{Viktor Popov}
\affiliation{Ascend Technologies Ltd, Southampton, United Kingdom}

\begin{abstract}
Dental aerosol procedures produce sub-\SI{50}{\micro\metre} nuclei that
can remain airborne for long periods in enclosed clinics, creating pathways
for airborne pathogen transmission.
Reynolds-Averaged Navier--Stokes (RANS) simulations with Euler--Lagrange
particle tracking capture this transport accurately but require very
long run times per scenario, which precludes real-time clinical
decision support in 3D.
We present the \emph{Eulerian--Lagrangian Graph Interaction Network}
(\ELGIN), a physics-informed graph surrogate that jointly
predicts carrier-flow dynamics on the OpenFOAM polyhedral mesh and the
per-parcel motion of the polydisperse spray cloud.
\ELGIN\ couples a multi-head Graph Transformer with Jacobi-preconditioned
learnable pressure projection and a turbulence-closure head to a
sigmoid-gated Lagrangian Interaction Network through differentiable
inverse-distance mesh--parcel coupling, and advances parcels with a
symplectic St\"ormer--Verlet integrator.
A four-stage physics-informed curriculum stabilises 260-step autoregressive
rollouts without gradient explosion.
A parameter sweep with foam-extend~4.1 OpenFOAM \texttt{reactingParcelFoam}
across clinically relevant ventilation rates and handpiece spray speeds
provides CFD ground truth.
This article reports a single-case demonstration in which both \ELGIN\ and
a Lagrangian-only baseline (M0) are trained and evaluated on
\texttt{Sweep\_Case\_03} of a twenty-case sweep; full
16/2/2 retraining is in progress and will replace all reported metrics.
On this case, \ELGIN\ tracks the foam-extend particle cloud much more
closely than M0: mean parcel
displacement error falls from \MzeroMDE\,\% to \elginMDE\,\% of room
width and cloud radius-of-gyration error from \MzeroRg\,\% to \elginRg\,\%.
A \SI{26}{\second} rollout completes in $\sim$\rolloutSec~s on a
\SI{4}{\giga\byte} GPU, approximately \speedupX$\times$ faster than the
foam-extend reference pipeline, toward per-appointment
infection-risk screening once the multi-case checkpoint is in place.
\end{abstract}

\maketitle

\section{\label{sec:intro}Introduction}

Dental procedures such as high-speed drilling
($\sim$\SI{300000}{rpm}), ultrasonic scaling, and air-polishing routinely
generate polydisperse bioaerosols with droplet nuclei in the
\SIrange{0.5}{50}{\micro\metre} range.\cite{Harrel2004,Micik1969}
These particles remain airborne for 30--90 minutes in poorly ventilated
clinical rooms\cite{Li2021} and carry bacteria
(\textit{M.\ tuberculosis}, oral streptococci), viruses
(SARS-CoV-2, hepatitis~B), and fungal spores that infect patients,
dental workers, and bystanders.\cite{Zemouri2017}
The COVID-19 pandemic dramatically reinvigorated concern about airborne
pathogen transmission in enclosed clinical
spaces,\cite{Morawska2020,Wang2021} motivating quantitative tools that
can predict where aerosols migrate under given ventilation conditions
and identify infection hot-zones in real time.

Computational Fluid Dynamics (CFD) has been the primary quantitative tool
for complex flow phenomena for over six decades.\cite{Ferziger2002}
For turbulent aerosol transport the Euler--Lagrange framework is widely
used: the carrier-phase turbulent flow is resolved by the
Reynolds-Averaged Navier--Stokes (RANS) equations with a $k$--$\omega$
SST (Shear Stress Transport) closure,\cite{Menter1994} while individual
particle trajectories are obtained by integrating Newton's second law
under aerodynamic drag (with Cunningham-corrected Stokes relaxation
time),\cite{Cunningham1910} gravity, turbulent dispersion via a Discrete
Random Walk (DRW) model,\cite{Pope2000} and Brownian
diffusion\cite{Einstein1905} for sub-micrometre nuclei.
Despite this physical fidelity, a single RANS~+~Lagrangian-parcel
simulation of a dental treatment room requires roughly
\SI{40}{\minute} of single-core wall time on a modern workstation in the
present foam-extend~4.1 reference pipeline (SI~Sec.~S7), rendering
Monte~Carlo sweeps over ventilation rates, room layouts, and spray angles
intractable within clinical timescales.
Published CFD studies of dental and indoor bioaerosol
transport\cite{Li2021,Zemouri2017,Gupta2010} have characterised the risk
qualitatively but none provides a tool capable of real-time personalised
guidance during a dental appointment.

The past decade has witnessed explosive growth in machine-learning
surrogates for CFD.
Convolutional Neural Networks (CNNs) on Cartesian grids,\cite{Guo2016}
embedded learned closures,\cite{Kochkov2021} Fourier Neural
Operators,\cite{LiFNO2021} DeepONet,\cite{Lu2021} and Physics-Informed
Neural Networks (PINNs)\cite{Raissi2019} all achieve $10^3$--$10^5\times$
speed-ups but share a fundamental limitation: they operate on
fixed-resolution structured grids or hand-crafted collocation points and
cannot generalise to the unstructured topology-varying meshes
characteristic of complex clinical geometries.
Graph Neural Networks (GNNs) overcome this limitation by representing the
simulation domain as a graph $\mathcal{G}=(\mathcal{V},\mathcal{E})$ in
which mesh cells or particles are nodes
$v_i\in\mathcal{V}$ with feature vectors
$\bm{h}_i\in\mathbb{R}^{d_v}$, edges $(i,j)\in\mathcal{E}$ carry edge
feature vectors $\bm{e}_{ij}$, and information propagates through $K$
rounds of \emph{message passing}:
\begin{align}
  \bm{m}_{ij}^{(k)} &= \phi_e^{(k)}\!\bigl(
    \bm{h}_i^{(k)},\,\bm{h}_j^{(k)},\,\bm{e}_{ij}\bigr),
  \label{eq:mp_msg}\\
  \bm{h}_i^{(k+1)} &= \phi_v^{(k)}\!\Bigl(
    \bm{h}_i^{(k)},\,\bigoplus_{j\in\mathcal{N}(i)}\bm{m}_{ij}^{(k)}\Bigr),
  \label{eq:mp_update}
\end{align}
where $\bm{m}_{ij}^{(k)}$ is the message passed from node $j$ to node~$i$
at round~$k$, $d_v$ is the dimension of $\bm{h}_i$, $\phi_e,\phi_v$ are learnable multi-layer perceptrons,
$\mathcal{N}(i)$ is the neighbourhood of node~$i$, and $\bigoplus$ is a
permutation-invariant aggregation operator.\cite{Gilmer2017}
This Message Passing Neural Network formalism is the formal generalisation
of finite-volume stencil updates to arbitrary graph topologies, and the
relational inductive bias\cite{Battaglia2018} aligns naturally with the
locality of physical interaction laws.

The Encode-Process-Decode (EPD) paradigm of Sanchez-Gonz\'{a}lez et
al.\cite{Sanchez2020} (the Graph Network Simulator, \GNS) and the
mesh-based MeshGraphNets of Pfaff et al.\cite{Pfaff2021} provide the
architectural foundation for most GNN-CFD surrogates.
For Lagrangian particle simulation the \GNS\ achieves rollout errors
below 5\% at $10^5\times$ the speed of Smoothed Particle Hydrodynamics
(SPH) and Material Point Method (MPM) solvers on granular benchmarks.
Recent extensions address ocean wave dynamics,\cite{Hanke2025} dense
granular suspensions,\cite{Aminimajd2025} plasma
particle-in-cell,\cite{Mlinarevik2025} SE(3)-equivariant momentum
conservation\cite{Sharma2025} (Special Euclidean group in three
dimensions), and probabilistic uncertainty quantification through
diffusion graph networks.\cite{Lino2025}
Physics-informed loss terms,\cite{Raissi2019} attention
aggregation,\cite{Velickovic2018,Brody2022} and Hamiltonian or Lagrangian
decoders\cite{Greydanus2019,Cranmer2020,Hairer2006} have all been
proposed to mitigate the persistent weakness of vanilla \GNS: purely
data-driven training does not guarantee energy or momentum
conservation, and rollouts drift over long horizons.\cite{Sharma2025}

Two concurrent works independently highlight the value of
Eulerian--Lagrangian hybridisation for GNN surrogates.
\textsc{corgi}\cite{CORGI2025} augments a GNS backbone with a
lightweight convolutional Eulerian grid for global context
aggregation, projecting particle features to a grid, applying CNN
updates, and mapping the result back to the particle domain, achieving
57\% better rollout accuracy over plain GNS with only 13\% inference
overhead.
\textsc{DeepLag}\cite{DeepLag2024} integrates Lagrangian particle
tracking into Eulerian field prediction via cross-attention
\textsc{EuLag} blocks, using tracked particles to guide the evolution
of the Eulerian velocity field.
Both methods use \emph{learned} Eulerian representations; neither
conditions on a pre-computed physical flow field, nor addresses
multi-physics particle dynamics.
For RANS-conditioned GNNs on Eulerian fields, \textsc{pignn-cfd}\cite{PIGNNCFD2023}
trains a physics-informed GNN on OpenFOAM RANS outputs to predict
urban wind fields on unstructured meshes at $10^2\times$ CFD speed,
demonstrating that RANS-mesh graphs are viable training inputs,
but does not track Lagrangian particles or handle particle physics.
In the aerosol domain, \textsc{glad}\cite{GLAD2025} applies the GNS
framework to \emph{atmospheric aerosol microphysics} (sulfuric acid
condensation and coagulation in the PartMC-MOSAIC model), demonstrating
GNS generalisation to particle chemical dynamics.
\textsc{glad} operates without a background flow field and targets
composition changes rather than spatial transport, leaving the
problem of turbulent indoor bioaerosol dispersion unaddressed.
A comprehensive comparison of GNN-CFD methods, methodologies, advantages,
and limitations is provided in the Supporting Information
(Sec.~S2, Table~S5).

A systematic examination of the GNN-CFD literature reveals that
existing Lagrangian GNN surrogates address dry granular
flows,\cite{Sanchez2020,Kumar2022} dense
suspensions,\cite{Aminimajd2025} free-surface waves,\cite{Hanke2025}
plasma particle-in-cell,\cite{Mlinarevik2025} or atmospheric aerosol
microphysics without spatial transport.\cite{GLAD2025}
RANS-conditioned GNNs have been demonstrated for Eulerian wind-field
prediction on urban OpenFOAM meshes,\cite{PIGNNCFD2023} and hybrid
Eulerian--Lagrangian architectures have been explored with learned
(rather than physically pre-computed) Eulerian
representations.\cite{CORGI2025,DeepLag2024}
No previous work reports a GNN surrogate for turbulent bioaerosol spatial
transport in enclosed clinical spaces that is conditioned on a
pre-computed physical carrier-phase flow field.
This gap is consequential, because dental aerosol physics in this regime
involves several effects that prior GNN benchmarks do not jointly cover.
Spray droplets are polydisperse (\SIrange{1}{50}{\micro\metre}), so
aerodynamic relaxation times span orders of magnitude with $d_p$.
The ventilation Stokes number $St$ (ratio of the Stokes relaxation time
at mean diameter $\bar{d}_p$ to $H/V_{\rm in}$, with
$H=\SI{3.0}{\metre}$ matching the CFD domain height; full definition
in SI~Sec.~S5) remains $[\StokesRange]$ across the sweep, i.e.\
tracer-like on the mean-flow time scale.
Sub-micrometre nuclei exhibit Cunningham slip-correction factors
$C_c>1.5$ that strongly modify drag.
Saffman shear-lift is significant for $d_p\approx\SI{3}{\micro\metre}$
and larger; Brownian diffusion dominates below
$d_p\approx\SI{0.3}{\micro\metre}$ and requires a stochastic model that
deterministic \GNS\ formulations omit.
The carrier flow is an Eulerian field tightly coupled to
obstacle-laden geometry (dentist, patient, walls, ceiling supply,
lateral pressure outlet), so dispersion depends on wall-normal proximity,
swirl, and recirculation.
Clinical deployment further requires calibrated risk intervals on
infection-relevant metrics.

This paper addresses these gaps through a hybrid Eulerian--Lagrangian
GNN surrogate framework, demonstrated here on the representative case
\texttt{Sweep\_Case\_03} of a twenty-case foam-extend~4.1
\texttt{reactingParcelFoam} parameter sweep of a two-dimensional
dental treatment room.
The present results are a \emph{single-case demonstration} that
illustrates the architecture, training curriculum and evaluation
pipeline on one realisation of the parameter grid;
the production checkpoint analysed below is trained and rolled out on
the same case, so the numbers should be read as proof-of-concept
fidelity rather than as a generalisation study.
Full 16/2/2 (train/validation/test) retraining on the complete
twenty-case dataset is in progress and results will be updated once the multi-case rollouts are available.

Two model variants are developed and compared: (i) a \GNS\ baseline (M0) following the Encode--Process--Decode paradigm of Sanchez-Gonz\'{a}lez
et al.,\cite{Sanchez2020} with uniform Lagrangian message aggregation
and no Eulerian carrier-field conditioning; and (ii) \ELGIN, the principal
contribution of this work, which solves
the carrier flow on the OpenFOAM polyMesh with a multi-head Graph
Transformer processor and a Jacobi-preconditioned learnable pressure
projection, replaces uniform aggregation with sigmoid-gated attention on
the particle graph,\cite{Velickovic2018} exposes every parcel node to the
projected RANS velocity,
the local turbulent kinetic energy, the obstacle-aware
distance-to-wall, and the wall-normal vector interpolated from the
mesh, broadcasts a per-case airInlet velocity vector as a global
conditioning feature, and integrates particle positions with a
symplectic St\"ormer--Verlet step.

The four primary contributions of this work are:
(i) a hybrid Eulerian--Lagrangian GNN surrogate for polydisperse dental
bioaerosol that jointly predicts the divergence-corrected RANS velocity
field and the particle trajectories on the same OpenFOAM polyMesh;
(ii) a full-timeline data-extraction protocol with persistent
\texttt{origId} tracking and an alive-mask training objective that
correctly handles variable parcel counts arising from injection and
deposition;
(iii) explicit, name-aware boundary-condition encoding (nine semantic
classes including dentist and patient obstacles), polyMesh-derived
distance-to-wall and wall-normal features, and a per-case airInlet
velocity conditioning vector designed to let a single trained model
generalise across ventilation rates and handpiece spray speeds (the
present single-case checkpoint exercises the conditioning mechanism;
demonstration of cross-case generalisation is deferred to the planned
16/2/2 retraining); and
(iv) a four-stage training curriculum (Eulerian pre-training, particle
supervised, PDE-informed joint, backpropagation-through-time rollout
fine-tuning) that stabilises long-horizon GNN rollouts for the
dissipative multi-physics dental problem.

The remainder of the paper is organised as follows.
Section~\ref{sec:cfd} summarises the CFD data-generation methodology
(detailed equations and solver settings are in SI~Sec.~S1).
Section~\ref{sec:gnn} presents the two GNN model variants.
Section~\ref{sec:training} outlines the four-stage training curriculum
(detailed loss formulation in SI~Sec.~S3).
Section~\ref{sec:results} reports headline rollout metrics, snapshot
comparisons, and the clinical Breathing Zone Exposure (BZE) outcome,
with extended results, non-dimensional analyses, dispersion statistics,
and computational benchmarks in SI~Secs.~S4--S7.
Section~\ref{sec:conclusion} concludes.

\section{\label{sec:cfd}CFD Data Generation}

\subsection{\label{subsec:geometry}Geometry, governing physics, and parameter sweep}

\begin{figure*}
\includegraphics[width=\textwidth]{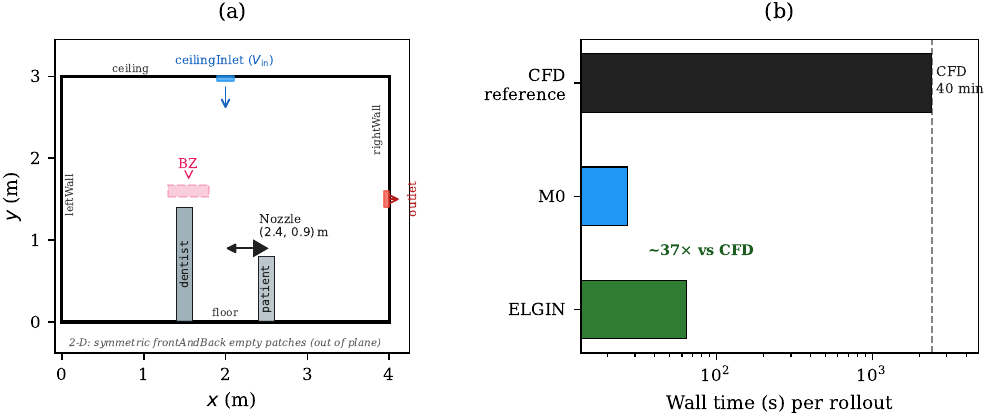}
\caption{\label{fig:framework}%
(a)~Dental treatment room: \SI{4}{\metre}~$\times$~\SI{3}{\metre}
2-D cross-section with every OpenFOAM boundary patch labelled
(\texttt{ceilingInlet}, \texttt{outlet}, \texttt{floor},
\texttt{ceiling}, \texttt{leftWall}, \texttt{rightWall},
\texttt{dentist}, \texttt{patient}, and symmetric
\texttt{frontAndBack} empty extrusion),
the dental-handpiece nozzle,
and the clinician breathing zone~(BZ).
(b)~Wall-clock time for one autoregressive surrogate rollout spanning the stored
\SI{26}{\second} trajectory segment versus the foam-extend
\texttt{reactingParcelFoam} reference pipeline for the same segment. }
\end{figure*}

The computational domain represents a simplified two-dimensional
cross-section of a dental treatment room
(\SI{4.0}{\metre}~$\times$~\SI{3.0}{\metre}, Fig.~\ref{fig:framework}a),
using Cartesian coordinates with vertical coordinate~$y$.
The two-dimensional cross-section is adopted to balance physical
representativeness with the computational cost of generating a
twenty-case sweep for GNN training. 
The domain contains a \SI{0.2}{\metre}-wide ceiling supply inlet
(downward velocity $V_{\rm in}$); a lateral pressure outlet at
$y = \SI{1.5}{\metre}$; rectangular obstacles representing the dentist
and patient; and a \SI{3}{\milli\metre} dental-handpiece nozzle
($d_{\rm nozzle}=\SI{3e-3}{\metre}$, in the typical range for
high-speed dental sprays) at the patient's oral cavity, aimed
horizontally toward the dentist with injection speed $U_{\rm mag}$
and cone half-angle $\theta$.
A breathing zone (BZ) is defined at the dentist's head height for
infection-risk metric evaluation; the Breathing Zone Exposure (BZE)
computed from this volume is specified in Sec.~\ref{subsec:bze}.
The full set of boundary conditions, mesh-independence study, and inlet
turbulence formulae are tabulated in SI~Sec.~S1.

The carrier-phase airflow is modelled by the incompressible RANS equations with the
Menter $k$--$\omega$ SST closure:\cite{Menter1994}
\begin{align}
  \nabla \cdot \vect{U} &= 0, \label{eq:cont}\\
  \frac{\partial \vect{U}}{\partial t}
  + \nabla \cdot (\vect{U}\vect{U}) &= -\nabla p
  + \nabla \cdot \left[(\nu + \nu_t)\nabla\vect{U}\right]
  + \vect{S}_p, \label{eq:mom}
\end{align}
where $\vect{U}$ is the Reynolds-averaged mean velocity,
$\nu$ and $\nu_t$ are the kinematic and turbulent eddy
viscosities, $p$ is the modified kinematic pressure, and $\vect{S}_p$ is
the particle back-reaction.
The eddy viscosity is closed by the SST formula and reported in full in
SI~Eq.~(S3).

Each parcel of instantaneous diameter $d_p(t)$ and mass $m_p$ obeys the
simplified
Maxey--Riley equation:\cite{Maxey1983}
\begin{equation}
  m_p \frac{d\vect{v}}{dt} = \vect{F}_{\rm drag} + \vect{F}_{\rm lift}
  + \vect{F}_g + \vect{F}_{\rm Br} + \vect{F}_{\rm turb},
  \label{eq:particle}
\end{equation}
where $\vect{v}$ is the parcel velocity,
$\vect{F}_{\rm drag}$ the Cunningham-corrected Stokes drag,
$\vect{F}_{\rm lift}$ the McLaughlin--Saffman shear-lift,
$\vect{F}_g = m_p \vect{g}$ the gravitational force with gravitational
acceleration~$\vect{g}$,
$\vect{F}_{\rm Br}$ the Brownian stochastic (thermal) force, and
$\vect{F}_{\rm turb}$ the turbulent dispersion force from the DRW model.
The full closed forms of each term,
the Cunningham slip-correction factor,
and the McLaughlin-corrected Saffman expression are given in
SI~Sec.~S1.3, and a force-magnitude regime map across
$d_p \in [0.1, 50]\,\mu$m is presented in SI~Sec.~S4.

Parcels are sampled from a Rosin--Rammler distribution
($d_{\min}=\SI{1}{\micro\metre}$, $d_{\max}=\SI{50}{\micro\metre}$) and
the foam-extend solver evaporates each parcel quasi-steadily according
to Wells' law,\cite{Wells1934}
\begin{equation}
  d_p(t)^2 = d_{p0}^2 - K\,t,
  \quad K = \frac{8\rho_{\rm air} D_v \ln(1 + B_M)}{\rho_p},
  \label{eq:wells}
\end{equation}
where $d_{p0}$ is the initial droplet diameter, $\rho_{\rm air}$ and
$\rho_p$ are the air and droplet densities, respectively,
$D_v = \SI{2.6e-5}{\metre\squared\per\second}$ is the vapour mass
diffusivity, and $B_M = 0.0263$ is the Spalding mass-transfer number
at \SI{50}{\percent} relative humidity.
The full Wells closure and its implementation in
\texttt{reactingParcelFoam} are documented in SI~Sec.~S1.3.5.
The training data are extracted at $\Delta t_{\rm save}=\SI{0.1}{\second}$
without further evaporation post-processing, and the GNN is configured
to treat the parcel diameter $d_p$ as a static per-parcel input
feature inherited from the initial Rosin--Rammler sampling
(diameter dynamics are not predicted by the surrogate).
This is consistent with the relatively small fractional change in $d_p$
over the typical advection time-scale resolved by the
$\Delta t_{\rm save}=\SI{0.1}{\second}$ snapshots, and avoids polluting
the learned dynamics with an additional time-scale that is not needed
for displacement and dispersion prediction; the Wells reduction in
inertia is implicitly absorbed into the trajectory targets generated
by the foam-extend solver.

Each case is integrated as a two-stage foam-extend~4.1 pipeline:
steady-state RANS via \texttt{simpleFoam}, followed by transient
compressible reacting parcel transport via \texttt{reactingParcelFoam}
for \SI{30}{\second} of physical time using PIMPLE (merged PISO--SIMPLE)
pressure--velocity coupling.
Twenty cases span a $4\times5$ factorial grid over
$V_{\rm in} \in \{0.10,\,0.20,\,0.35,\,0.50\}\,\mathrm{m\,s^{-1}}$ and
$U_{\rm mag} \in \{10,\,20,\,30,\,40,\,50\}\,\mathrm{m\,s^{-1}}$ at fixed
$\theta = 20^\circ$, providing 261 quasi-steady-state frames per case
(extracted on $t \in [\SI{2}{\second},\SI{28}{\second}]$ at
$\Delta t_{\rm save} = \SI{0.1}{\second}$) split 16/2/2
(train/validation/test) by random shuffle with a fixed seed.
The OpenFOAM polyMesh contains $N_c = 7704$ owner cells with nine
named boundary patches (\texttt{ceilingInlet}, \texttt{outlet},
\texttt{floor}, \texttt{ceiling}, \texttt{leftWall}, \texttt{rightWall},
\texttt{dentist}, \texttt{patient}, and the \texttt{frontAndBack} empty
patch enforced by the 2-D extrusion).
Figure~\ref{fig:framework}(a) sketches the associated patch layout,
nozzle, and breathing zone.
Detailed solver settings, the full case list, and dataset statistics are
provided in SI~Tables~S2--S4.
For each timestep, $N_{\rm sub} = 1000$ representative parcel
trajectories are tracked with persistent \texttt{origId} identifiers so
that every node in the Lagrangian graph corresponds to the same physical
parcel across all frames (deposited or out-of-domain parcels are masked
out by an alive-mask flag rather than re-indexed).

\section{\label{sec:gnn}GNN Architectures}

The \ELGIN\ surrogate operates on a \emph{dual-graph} representation:
a static \emph{Eulerian mesh graph}
$\mathcal{G}^E=(\mathcal{V}^E,\mathcal{E}^E)$ derived from the
foam-extend polyMesh, and a time-varying \emph{Lagrangian particle graph}
$\mathcal{G}^L_t=(\mathcal{V}^L_t,\mathcal{E}^L_t)$ rebuilt at every
rollout step.
The two graphs are coupled through a sparse cross-graph
inverse-distance-weighted (IDW) interpolation operator
(Sec.~\ref{subsec:graph}).
The two model variants compared in this paper~--~\GNS\ (M0) and \ELGIN~--~differ by the carrier-phase information
they consume, by Lagrangian message aggregation (uniform sum vs.\ sigmoid-gated attention), and by the position integrator, but
share the same Encode--Process--Decode (EPD)
backbone.\cite{Battaglia2018}
The Lagrangian side employs a long short-term memory (LSTM)\cite{Hochreiter1997}
velocity-history encoder and an optional variational autoencoder (VAE)-style
acceleration decoder (Sec.~\ref{subsec:lagrangian}).
The full \ELGIN\ is described first
(Secs.~\ref{subsec:graph}--\ref{subsec:lagrangian});
M0 is obtained by switching off specific feature flags
(Sec.~\ref{subsec:variants}).

\begin{figure*}
\includegraphics[width=\textwidth]{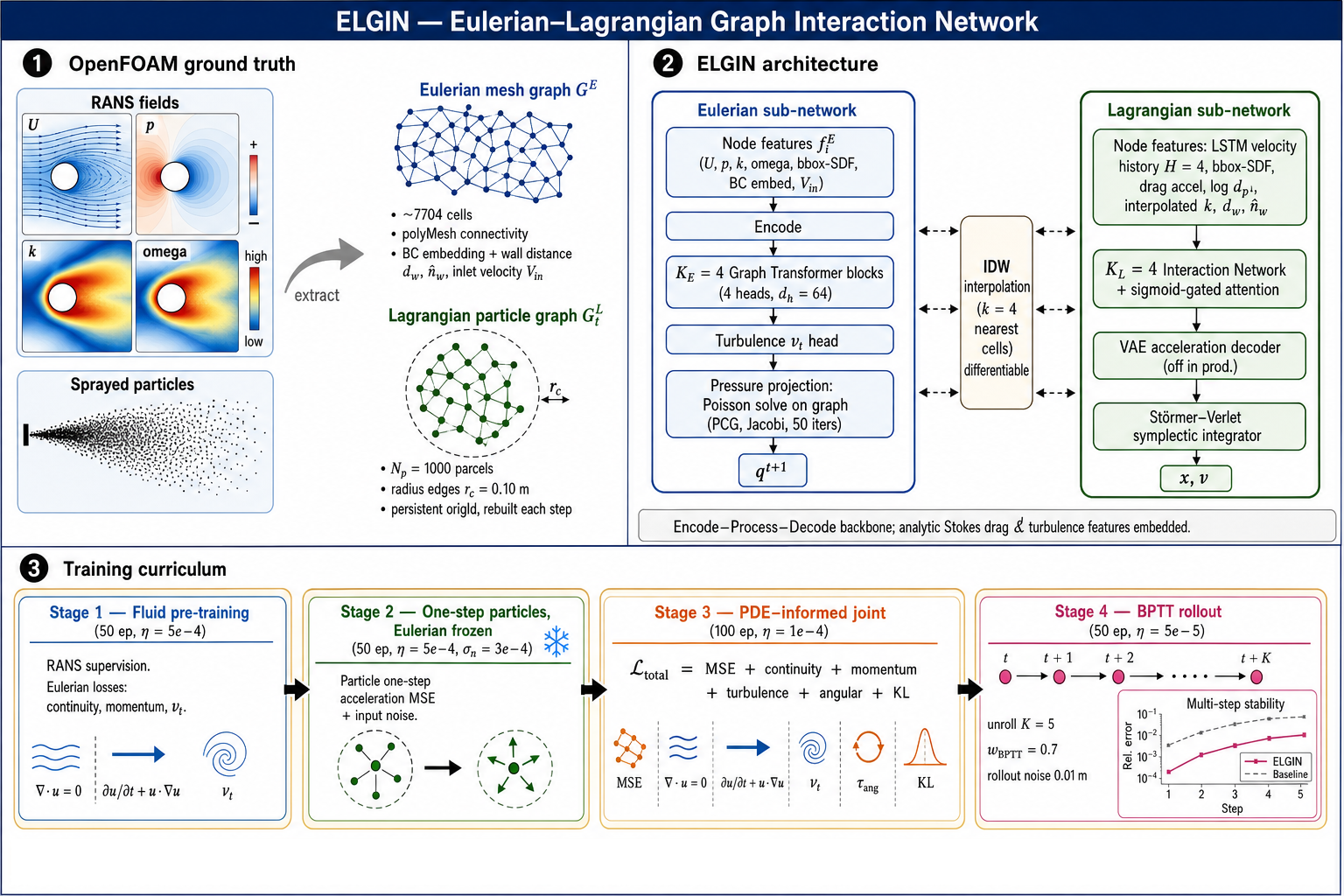}
\caption{\label{fig:elgin_arch}%
Overview of the \ELGIN\ workflow: dual-graph extraction from OpenFOAM
RANS and parcel data, coupled Eulerian and Lagrangian sub-networks with
differentiable inverse-distance exchange, and the four-stage training
curriculum (the same backbone and schedule apply to the M0 baseline).
Full notation, loss terms, and equation references are given in
Sec.~\ref{sec:gnn}.}
\end{figure*}

Figure~\ref{fig:elgin_arch} gives an end-to-end schematic of the
\ELGIN\ pipeline.
\emph{(Panel~1)} depicts the OpenFOAM ground truth: RANS fields
$\vect{U}$, $p$, $k$, and $\omega$ on the polyMesh together with the
sprayed Lagrangian parcels.
The dual-graph representation extracted from this data couples a
static Eulerian mesh graph $\mathcal{G}^E$ on $\sim$\,7\,700 cells,
equipped with boundary-condition embeddings, distance-to-wall $d_w$,
wall-normal $\hat{\vect{n}}_{\rm w}$, and the inlet-velocity vector
$\vect{V}_{\rm in}$, to a time-varying Lagrangian radius graph
$\mathcal{G}^L_t$ over
$N_p=1000$ persistent-\texttt{origId} parcels with connectivity radius
$r_c=\SI{0.10}{\metre}$.
\emph{(Panel~2)} shows the dual sub-network architecture.
The Eulerian branch is an Encode--Process--Decode stack of Graph
Transformer blocks based on the masked-attention dot-product
formulation of Shi et al.\cite{Shi2021}, followed by a
turbulence-closure head and a Jacobi-preconditioned learnable pressure
projection (Eq.~\ref{eq:proj}).
The Lagrangian branch combines an LSTM velocity-history encoder, an
Interaction-Network processor with sigmoid-gated attention
(Eq.~\ref{eq:attn_score}), an optional VAE-style acceleration decoder
(Sec.~\ref{subsec:lagrangian}), and a
symplectic St\"ormer--Verlet integrator (Eq.~\ref{eq:verlet}).
Information is exchanged through a differentiable inverse-distance-weighted
operator with $k_{\rm IDW}=4$ nearest cells (Eq.~\ref{eq:idw}).
\emph{(Panel~3)} outlines the four-stage curriculum used for both M0 and
\ELGIN.
Stage~1 is Eulerian fluid pre-training.
Stage~2 is one-step particle supervision with input noise.
Stage~3 is PDE-informed joint training with continuity, momentum,
turbulence, angular, and Kullback--Leibler (KL) terms.
Stage~4 is back-propagation through time (BPTT) rollout fine-tuning.
The architectural and training hyperparameter ($K_E$, $K_L$, $d_h$, batch size, per-stage epoch counts, learning rates, BPTT unroll length, noise scales) are provided in Table~\ref{tab:training}.

\begin{table}
\caption{\label{tab:training}Training hyperparameters common to both
model variants (20-case run).}
\begin{ruledtabular}
\begin{tabular}{ll}
Parameter & Value \\
\hline
Optimiser                  & AdamW,\cite{Kingma2015} weight decay $10^{-5}$ \\
LR schedule                & Cosine annealing ($\eta_{\rm min} = 0.01\,\eta_0$) \\
Total epochs               & 300 (both variants) \\
Stage allocation           & \ELGIN: 60/60/120/60; M0: 75/150/75 (Stage~1 skipped) \\
Batch size                 & 8 \\
Hidden dimension $d_h$     & 64 \\
Message-passing steps      & $K_L=4$ (Lagrangian); $K_E=4$ (Eulerian) \\
Lagrangian connectivity $r_c$ & \SI{0.10}{\metre} (\ELGIN); \SI{0.30}{\metre} (M0 baseline) \\
Cross-graph IDW neighbours $k_{\rm IDW}$ & 4 \\
Pressure-projection PCG iters & 20 (Stages 1--2); 50 (Stages 3--4); Jacobi-precond. \\
History length $H$         & 5 positions (= 4 velocity diffs); LSTM-encoded in \ELGIN \\
Model time step $\Delta t$ & \SI{0.1}{\second} (= CFD snapshot interval $\Delta t_{\rm save}$) \\
Lagrangian integrator      & St\"ormer--Verlet (\ELGIN); Euler (M0) \\
Position-noise scale       & $\sigma_n = 3\times10^{-4}$ (isotropic Gaussian) \\
BPTT unroll steps          & 5 (Stage~4) \\
BPTT loss weight $w_{\rm BPTT}$ & 0.7 \\
BPTT rollout noise         & $\sigma_{\rm roll}=0.01$~m \\
Stochastic decoder KL weight $\lambda_{\rm KL}$ & $10^{-3}$ (decoder off in production checkpoint) \\
GPU                        & NVIDIA Quadro P1000 (\SI{4}{\giga\byte} VRAM) \\
\end{tabular}
\end{ruledtabular}
\end{table}

\subsection{\label{subsec:graph}Dual-graph construction}

The polyMesh of $N_c=7704$ owner cells defines $|\mathcal{V}^E|=N_c$
nodes (one per cell centroid) and $|\mathcal{E}^E|$ undirected edges
through the internal face connectivity.
At every internal face, the owner--neighbour pair is added in both
directions, giving a fully antisymmetric processor.
The polyMesh boundary patches are parsed by a name-aware
classifier into nine semantic classes:
\texttt{interior}, \texttt{inlet} (\texttt{ceilingInlet}),
\texttt{outlet}, \texttt{wall} (\texttt{leftWall}, \texttt{rightWall}),
\texttt{floor}, \texttt{ceiling}, \texttt{dentist}, \texttt{patient},
and the \texttt{symmetryPlane}/\texttt{empty} 2-D extrusion class.
Each cell adjacent to a non-interior, non-empty boundary face inherits
the corresponding boundary-condition (BC) identifier
(\emph{first-wins} so that the front/back \texttt{empty} patch never
overwrites a physical wall, dentist, patient, inlet or outlet).
For each cell we compute the true geometric distance to the nearest
\emph{wall-type} face (walls + dentist + patient) directly from the
polyMesh,
\begin{equation}
  d_w^{(i)} \;=\; \min_{f\in\mathcal{F}_{\rm wall}}
    \bigl\lVert\vect{x}^{(i)}_{\rm cell} - \vect{x}^{(f)}_{\rm face}
    \bigr\rVert,
  \label{eq:dwall}
\end{equation}
where $\mathcal{F}_{\rm wall}$ is the set of obstacle and wall boundary
faces (walls, floor, ceiling, dentist, patient),
$\vect{x}^{(i)}_{\rm cell}$ is the centroid of cell~$i$, and
$\vect{x}^{(f)}_{\rm face}$ is the midpoint of face~$f$,
together with the unit wall-normal vector $\hat{\vect{n}}^{(i)}_{\rm w}$
pointing from the cell centroid towards the closest wall-face midpoint.
The polyMesh axis-aligned bounding box defines the
\emph{domain bounds} $[\vect{x}_{\min},\vect{x}_{\max}]$ that constrain
the rollout (Sec.~\ref{subsec:lagrangian}).

Each Eulerian node $i\in\mathcal{V}^E$ carries the feature vector
\begin{equation}
  \vect{f}_i^{E} = \bigl[\widetilde{\vect{q}}_i,\;
                          \widetilde{\vect{x}}_i^{\rm bbox},\;
                          \mathrm{Embed}_{\rm bc}(b_i),\;
                          \widetilde{\vect{V}}_{\rm in}\bigr],
  \label{eq:euler_node_feat}
\end{equation}
where
$\widetilde{\vect{q}}_i = (\widetilde{U}_x,\widetilde{U}_y,
\widetilde{p},\widetilde{k},\widetilde{\omega})_i$ is the per-channel
$z$-score of the five RANS state variables;
$\widetilde{\vect{x}}_i^{\rm bbox}\in\mathbb{R}^{2D}$ is the
two-sided signed distance to the four bounding-box walls;
$\mathrm{Embed}_{\rm bc}:\{0,\dots,15\}\!\to\!\mathbb{R}^{8}$ is a
learned look-up table over BC identifiers $b_i$; and
$\widetilde{\vect{V}}_{\rm in}\!\in\!\mathbb{R}^{2}$ is the per-case
\emph{airInlet velocity vector} parsed directly from the OpenFOAM
\texttt{0/U} dictionary, broadcast to every node so that a single
trained model generalises across the ventilation grid.
Eulerian edges concatenate a six-dimensional geometric descriptor (the
outward face unit normal $(\hat n_x,\hat n_y)$, the face area scaled by
$L_{\rm ref}^{-2}$, the Euclidean distance $\|\Delta\vect{x}_{ij}\|$
between owner and neighbour cell centroids scaled by $L_{\rm ref}^{-1}$,
and the unit separation vector
$\Delta\vect{x}_{ij}/\|\Delta\vect{x}_{ij}\|$ in two components) with a
four-dimensional learned face-type embedding,
giving $d_{\rm in}^{E,\rm edge}=10$.
The length $L_{\rm ref}=\SI{4.0}{\metre}$ is the nominal room width and
matches the reference scale used for trajectory-error metrics in
Sec.~\ref{subsec:metrics}.

At each timestep $t$, the $N_p=N_{\rm sub}=1000$ tracked parcels
(uniquely identified by their persistent foam-extend \texttt{origId})
form a directed radius graph,
\begin{equation}
  \mathcal{E}^L_t = \bigl\{(i,j):\|\vect{x}_i^{(t)} - \vect{x}_j^{(t)}\|
  < r_c,\; i\neq j\bigr\},
  \quad r_c = \SI{0.10}{\metre},
  \label{eq:edges}
\end{equation}
where $\vect{x}_i^{(t)}$ is the position of parcel $i$ at time~$t$.
Both orientations $(i,j)$ and $(j,i)$ are retained to preserve antisymmetry under
relative velocity.
The connectivity radius is set by the median nearest-neighbour spacing
in the spray core to keep the mean node degree below~20 while still
resolving aerodynamic wake interactions, and avoids the over-squashing
pathology of dense graphs.\cite{Battaglia2018}

Each Lagrangian node concatenates an LSTM-encoded velocity history,
geometric box-SDF features, a learned particle-type embedding, the
analytic Cunningham-corrected Stokes-drag acceleration, the log
diameter, the local turbulent kinetic energy interpolated from
$\mathcal{G}^E$, and~--~importantly~--~the \emph{cell-interpolated}
distance-to-wall and unit wall-normal pulled from
Eq.~(\ref{eq:dwall}).
Concretely,
\begin{equation}
  \vect{f}_i^{L} = \bigl[\,
    \mathrm{LSTM}\!\bigl(\widetilde{\vect{v}}_i^{(t-H:t)}\bigr),\;
    \widetilde{\vect{x}}_i^{\rm bbox},\;
    \mathrm{Embed}_{\rm type}(\tau_i),\;
    \widetilde{\vect{a}}^{\rm drag}_i,\;
    \log\widetilde{d}_p^{(i)},\;
    \widetilde{k}_i,\;
    \widetilde{d}_w^{(i)},\;
    \hat{\vect{n}}^{(i)}_{\rm w}
  \bigr],
  \label{eq:lag_node_feat}
\end{equation}
where the LSTM history encoder embeds the $H=4$ most recent
finite-difference velocities into a $d_{\ell}=32$-dimensional state,
$\tau_i$~is a discrete particle-type label feeding
$\mathrm{Embed}_{\rm type}$, and tildes mark training-set-normalised channels.
Edges in $\mathcal{G}^L_t$ use a local-frame geometric representation
whose first three components are invariant under planar rotations and
translations of the spray cone, while the fourth (the analytic drag
acceleration) is retained in the global Cartesian frame:
\begin{equation}
  \vect{f}_{ij}^{L,\rm edge} = \bigl[
    \log(1{+}\rho_{ij}),\,
    \cos\theta_{ij},\,
    \sin\theta_{ij},\,
    \widetilde{\vect{a}}^{\rm drag}_i
  \bigr],
  \label{eq:equiv_edge}
\end{equation}
with $\rho_{ij}=\|\vect{x}_j-\vect{x}_i\|/r_c$ and $\theta_{ij}$ the
relative bearing in the source-particle local frame.
The composition is thus rotation-\emph{invariant} in the geometric
descriptor and rotation-\emph{covariant} in the drag channel; full
SE(2)-equivariance would require expressing
$\widetilde{\vect{a}}^{\rm drag}_i$ in the same local frame, which is a
straightforward extension reserved for the planned full
twenty-case retraining.

Eulerian-to-Lagrangian feature transfer at any continuous parcel
position $\vect{x}_p$ uses an inverse-distance-weighted interpolation
over the $k_{\rm IDW}=4$ nearest cell centroids,
\begin{equation}
  \widetilde{\Phi}(\vect{x}_p) = \sum_{i\in\mathcal{N}_k(\vect{x}_p)}
    \frac{w_{p,i}\,\Phi_i}{\sum_{i'} w_{p,i'}},
  \quad
  w_{p,i} = \frac{1}{\|\vect{x}_p-\vect{x}_i\|+\varepsilon},
  \label{eq:idw}
\end{equation}
where $\Phi$ denotes any scalar field component carried on the mesh,
$\Phi_i$ is its value at cell~$i$, $\vect{x}_i$ is the centroid of
cell~$i$, $\mathcal{N}_k(\vect{x}_p)$ is the set of the
$k{=}k_{\rm IDW}$ nearest cell centroids to $\vect{x}_p$, and
$\varepsilon$ is a small positive constant for numerical
stability,
which is differentiable and respects the polyMesh's unstructured
topology (no rectangular bilinear assumption).
The same operator transports $d_w$ and $\hat{\vect{n}}_{\rm w}$ from
cells to parcels for the Lagrangian wall feature in
Eq.~(\ref{eq:lag_node_feat}).

\subsection{\label{subsec:eulerian}Eulerian sub-network and pressure projection}

The Eulerian sub-network advances the carrier flow on
$\mathcal{G}^E$ as
\begin{equation}
  \widehat{\vect{q}}^{(t+1)} \;=\; \vect{q}^{(t)} +
  \widehat{\Delta\vect{q}}^{(t)}_\theta,
  \label{eq:euler_step}
\end{equation}
where $\vect{q}^{(t)}$ stacks the five RANS fields
$(U_x,U_y,p,k,\omega)$ on each mesh cell at training step~$t$ and
$\widehat{\Delta\vect{q}}^{(t)}_\theta$ is the predicted residual increment
from the network with parameters~$\theta$.
The increment is produced by an EPD network whose processor is a stack of $K_E=4$
multi-head graph-transformer blocks with $H_{\rm att}=4$ heads each;
node updates use small multi-layer perceptron (MLP) maps denoted $\mathrm{MLP}_n^{(k)}$ below.
\begin{align}
  \alpha_{ij}^{(k,h)} &= \mathrm{softmax}_{j\in\mathcal{N}(i)}\!
     \frac{\bigl(\vect{W}_Q^{(k,h)}\vect{h}_i^{(k)}\bigr)^{\!\top}
           \bigl(\vect{W}_K^{(k,h)}\vect{h}_j^{(k)}+\vect{W}_E^{(k,h)}\vect{e}_{ij}^{(k)}\bigr)}
          {\sqrt{d_h/H_{\rm att}}},
  \label{eq:gtrans_attn}\\
  \vect{h}_i^{(k+1)} &= \vect{h}_i^{(k)} + \mathrm{MLP}_n^{(k)}\!\Bigl(
    \big\|_{h=1}^{H_{\rm att}} \!\sum_{j\in\mathcal{N}(i)}
      \alpha_{ij}^{(k,h)}\,\vect{W}_V^{(k,h)}\vect{h}_j^{(k)}\Bigr),
  \label{eq:gtrans_update}
\end{align}
with hidden dimension $d_h=64$, residual connections and layer
normalisation.
Here $\vect{h}_i^{(k)}$ is the Eulerian node embedding,
$\vect{e}_{ij}^{(k)}$ the edge feature on the owner--neighbour link
$(i,j)$ on $\mathcal{G}^E$,
$\vect{W}_Q^{(k,h)},\,\vect{W}_K^{(k,h)},\,\vect{W}_V^{(k,h)}$, and
$\vect{W}_E^{(k,h)}$ are learnable attention projections for head~$h$,
$\alpha_{ij}^{(k,h)}$ are attention weights, the concatenation
$\lVert_{h=1}^{H_{\rm att}}$ stacks head outputs, and $\mathcal{N}(i)$
is the set of graph neighbours of node~$i$ on $\mathcal{G}^E$.
This is the canonical \emph{softmax} dot-product attention of
graph-transformer architectures\cite{Shi2021,Vaswani2017} adapted to
the message-passing setting of Battaglia et al.\cite{Battaglia2018}, and
is consequently distinct from the sigmoid-gated graph attention
network (GAT) used on the Lagrangian side
(Sec.~\ref{subsec:lagrangian}).

A learnable \emph{turbulence closure} head
$\widehat{\nu}_t = \mathrm{MLP}_{\nu}(\vect{h}_i^{(K_E)})$ provides an
explicit eddy-viscosity prediction at every cell, regularised towards
the algebraic SST formula (Eq.~S3) so that the network does not have to
re-discover the closure from scratch.
A learnable \emph{pressure-projection} module reduces the discrete
divergence of the predicted velocity update by solving a graph-Poisson
problem
\begin{equation}
  \nabla_h \!\cdot\!\bigl(\nabla_h \phi\bigr) \;=\; \nabla_h\!\cdot\!
  \widehat{\vect{U}}^{(t+1)},\qquad
  \widehat{\vect{U}}^{\rm proj} \;=\; \widehat{\vect{U}}^{(t+1)}
  - \nabla_h\phi,
  \label{eq:proj}
\end{equation}
where $\phi$ is a scalar pressure-correction potential on the mesh graph
and $\nabla_h$ is a graph-edge approximation of the finite-volume gradient
operator (it uses the same edge set as the message-passing processor;
it is not the OpenFOAM \texttt{div(phi)} face-flux operator with
non-orthogonal corrections).
The projection therefore enforces zero net signed-edge flux on the graph
stencil and is consistent with, but not identical to, an exact
finite-volume divergence-free constraint.
The discrete graph-Laplacian is inverted by a fixed number of
Jacobi-preconditioned conjugate-gradient (PCG) iterations on
$\mathcal{G}^E$; the inner solve is fully differentiable and adds no
mesh-structure assumption beyond what the graph already encodes.
The combined Eulerian sub-network (encoder, $K_E$ graph-transformer
blocks, decoder, turbulence head, pressure projection) is pre-trained
in Stage~1 of the curriculum (Sec.~\ref{sec:training}) and held
nearly frozen thereafter.

\subsection{\label{subsec:lagrangian}Lagrangian sub-network}

The Lagrangian processor is a stack of $K_L=4$ Interaction Network
blocks with the same encode--process--decode skeleton as the Eulerian
side but operating on $\mathcal{G}^L_t$:
\begin{align}
  \vect{m}_{ij}^{(k)} &= \mathrm{MLP}_e^{(k)}\!\bigl(
    \vect{h}_i^{(k)}\,\|\,\vect{h}_j^{(k)}\,\|\,\vect{e}_{ij}^{(k)}\bigr),
  \label{eq:lag_msg}\\
  \bar{\vect{m}}_i^{(k)} &= \!\!\sum_{j\in\mathcal{N}_L(i)}\!
      \alpha_{ij}^{(k)}\,\vect{m}_{ij}^{(k)},
  \label{eq:lag_agg}\\
  \vect{h}_i^{(k+1)} &= \vect{h}_i^{(k)} + \mathrm{MLP}_n^{(k)}\!\Bigl(
    \vect{h}_i^{(k)}\,\|\,\bar{\vect{m}}_i^{(k)}\Bigr),
  \label{eq:lag_update}
\end{align}
where $\vect{m}_{ij}^{(k)}$ is the Lagrangian message from parcel $j$
to parcel~$i$, $\bar{\vect{m}}_i^{(k)}$ the aggregated message,
$\vect{h}_i^{(k)}$ the parcel node embedding,
$\mathcal{N}_L(i)$ is the set of Lagrangian neighbours of
parcel~$i$ under Eq.~(\ref{eq:edges}), $\alpha_{ij}^{(k)}\!\in\!\{1,\sigma(e_{ij}^{(k)})\}$ selects
between uniform sum aggregation and sigmoid-gated attention,
controlled by the model variant (Sec.~\ref{subsec:variants});
$e_{ij}^{(k)}$ is the scalar gate logit produced from the Lagrangian
edge encoding $\vect{e}_{ij}^{(k)}$.
An optional \emph{probabilistic (VAE-style) decoder}~\cite{Kingma2014}
maps the final node latent state to a normalised acceleration
$\widehat{\vect{a}}_i\in\mathbb{R}^{2}$ via a $\mu$/$\log\sigma^2$ head pair
and the reparametrisation trick, with the associated KL divergence
$\mathrm{KL}(\mathcal{N}(\mu,\sigma)\|\mathcal{N}(\vect{0},\vect{I}))$
added to Eq.~(\ref{eq:total_loss}) with weight
$\lambda_{\rm KL}=10^{-3}$.

Particle positions are advanced with a symplectic St\"ormer--Verlet
kick--drift--kick scheme,\cite{Hairer2006}
\begin{equation}
\begin{aligned}
  \vect{v}_i^{(t+1/2)} &= \vect{v}_i^{(t)} +
    \tfrac{1}{2}\Delta t\,\sigma_a\,\widehat{\vect{a}}_i^{(t)},\\
  \vect{x}_i^{(t+1)}   &= \vect{x}_i^{(t)} +
    \Delta t\,\vect{v}_i^{(t+1/2)},\\
  \vect{v}_i^{(t+1)}   &= \vect{v}_i^{(t+1/2)} +
    \tfrac{1}{2}\Delta t\,\sigma_a\,\widehat{\vect{a}}_i^{(t+1)},
\end{aligned}
\label{eq:verlet}
\end{equation}
Here $\sigma_a$ scales decoded accelerations into physical units
consistent with the ground-truth training-set statistics; 
$\vect{x}_i^{(t)}$ and $\vect{v}_i^{(t)}$ are parcel position and velocity.
The integration preserves a discrete symplectic two-form\cite{Hairer2006} and prevents the
energy drift typical of forward-Euler GNS rollouts on conservative
flows.\cite{Sanchez2020}
Predicted positions are clipped to the polyMesh
$[\vect{x}_{\min},\vect{x}_{\max}]$ bounds at every step, and a
\emph{geometry-aware deposition} flag is raised whenever
$\min(d_w^{(i,\rm IDW)},\,\delta_{\rm bbox}^{(i)})$ falls below an
adhesive threshold, where $\delta_{\rm bbox}^{(i)}$ is the shortest
distance from parcel~$i$ to the rectangular domain bbox;
once raised, the parcel's velocity is set to zero
for the remainder of the rollout, mimicking the foam-extend
\texttt{stick} wall interaction.

The Lagrangian network is designed so that closed-form particle physics
need not be learned from scratch: at every step the analytic
Cunningham-corrected Stokes drag and the SST turbulent-kinetic-energy
field enter the node and edge features
(Eq.~\ref{eq:lag_node_feat}, \ref{eq:equiv_edge}); a
discrete-random-walk (DRW) eddy-interaction model adds an analytic
turbulent kick scaled by $\sqrt{2k/3}$.

\subsection{\label{subsec:variants}Two model variants}

The two GNN surrogates compared in this work are configurations of
the same hybrid backbone, summarised in Table~\ref{tab:gns_compare}:
\begin{itemize}
\item \GNS\ (M0): Lagrangian-only baseline with sum aggregation, a
flat velocity-history input (no LSTM), simple relative-position edges,
forward-Euler integration, and \emph{no} carrier-flow conditioning.
This is a faithful reproduction of the canonical GNS of
Sanchez-Gonz\'{a}lez et al.\cite{Sanchez2020}
The M0 connectivity radius is set to $r_c^{(M0)}=\SI{0.30}{\metre}$,
larger than the ELGIN value below, so that the purely Lagrangian
neighbourhood reaches across the spray cone and partially compensates
for the absence of an Eulerian context; reducing $r_c^{(M0)}$ to the
\SI{0.10}{\metre} ELGIN value gave noticeably worse M0 validation MDE
in preliminary tests because parcels lose neighbours as the cloud
disperses. This choice is therefore conservative in favour of M0.
\item \ELGIN: the full hybrid model.  Lagrangian message aggregation
uses sigmoid-gated graph attention (Eq.~\ref{eq:attn_score}) rather than uniform
sums, consistent with Veli\v{c}kovi\'{c} et al.\cite{Velickovic2018};
the Eulerian sub-network of Sec.~\ref{subsec:eulerian} runs at every
step, the Lagrangian processor consumes the projected RANS velocity
through the analytic drag/Saffman/turbulent kinetic energy (TKE) features, the LSTM history
encoder (Eq.~\ref{eq:lag_node_feat}) and rotation-invariant local-frame
edges (Eq.~\ref{eq:equiv_edge}) replace the minimal M0 inputs, the
distance-to-wall and wall-normal interpolated from the polyMesh enter
both the Eulerian and the Lagrangian features, the per-case airInlet
vector is broadcast as global conditioning, and the symplectic
St\"ormer--Verlet integrator of Eq.~(\ref{eq:verlet}) replaces forward
Euler.
A tighter Lagrangian radius $r_c^{(\rm ELGIN)}=\SI{0.10}{\metre}$
suffices here because the cell-interpolated RANS velocity already
supplies the long-range carrier information, so the network does not
need a wider parcel neighbourhood to recover it.
\end{itemize}
The attention gate in \ELGIN\ takes the form
\begin{equation}
  \alpha_{ij}^{(k)} = \sigma\!\Bigl(
    \vect{a}^{(k)\top}\mathrm{LeakyReLU}\!\bigl[
      \vect{W}^{(k)}\vect{h}_i^{(k)}\,\|\,
      \vect{W}^{(k)}\vect{h}_j^{(k)}\,\|\,
      \vect{e}_{ij}^{(k)}\bigr]\Bigr),
  \label{eq:attn_score}
\end{equation}
with learned row vector $\vect{a}^{(k)}$ and weight matrices
$\vect{W}^{(k)}$, and $\|$ denoting feature concatenation.
The sigmoid (rather than softmax) yields an absolute relevance score
in $(0,1)$ and lets the model down-weight distant neighbours; in the
spray core the weights concentrate on the nearest fluid-coupled
parcels, recovering an implicit Oseen-type distance
decay.\cite{Oseen1910}
Crucially, \emph{neither model predicts a per-step diameter
change}: $d_p$ enters Eq.~(\ref{eq:lag_node_feat}) as a static
per-parcel input feature inherited from the Rosin--Rammler initial
distribution, and the surrogate is a position--velocity predictor
only.
This is a deliberate design choice motivated by two observations.
First, sub-\SI{10}{\micro\metre} droplets dominate clinical aerosol
exposure risk, and for that fraction the per-step fractional change
in $d_p^2$ is below \SI{1}{\percent} (SI~Sec.~S1.3.5), so the static-$d_p$
approximation is essentially exact on the
$\Delta t_{\rm save}=\SI{0.1}{\second}$ snapshot grid.
Second, the foam-extend trajectory targets already absorb the
$\tau_p\propto d_p^2$ inertia reduction caused by Wells' law (with full
\SI{50}{\micro\metre} droplets evaporating to their nuclei in
$\sim\SI{0.4}{\second}$, SI~Eq.~(S20)), so the GNN learns a
position--velocity mapping that implicitly inherits the diameter drift
from the target data without having to track it explicitly.
Diameter dynamics, if required by a downstream infection-risk
calculation, can be applied as an analytic post-processing step on the
predicted trajectories.

\begin{table}
\caption{\label{tab:gns_compare}%
Architecture comparison for the two GNN surrogates.
$K_L$~is the number of Lagrangian message-passing blocks; the Eulerian
sub-network and pressure projection are only active in \ELGIN.}
\begin{ruledtabular}
\begin{tabular}{lll}
Model & Lagrangian aggregation & Carrier-phase conditioning \\
\hline
\GNS\ (M0)     & Uniform sum       & none \\
\ELGIN\       & Sigmoid attention & full Eulerian + pressure proj. \\
\end{tabular}
\end{ruledtabular}
\end{table}

\section{\label{sec:training}Training Protocol}

Both model variants share the same four-stage curriculum (panel~3 of
Fig.~\ref{fig:elgin_arch}).
The AdamW optimiser\cite{Kingma2015} with weight decay $10^{-5}$ and
cosine-annealing learning-rate schedule is used throughout, on an
NVIDIA Quadro P1000 GPU (\SI{4}{\giga\byte} VRAM) using
PyTorch~2.x\cite{Paszke2019} and PyTorch Geometric.\cite{Fey2019}
The full epoch budget of 300 epochs is distributed across the four
stages as 60\,:\,60\,:\,120\,:\,60, allocating the bulk of the budget to
the PDE-informed joint Stage~3 and the BPTT rollout fine-tuning
Stage~4:
The production checkpoint analysed in Sec.~\ref{sec:results} is trained on the single representative case \texttt{Sweep\_Case\_03} of the twenty-case CFD design space.

\textbf{Stage~1 (Eulerian fluid pre-training, 60 epochs,
$\eta_0 = 5\times10^{-4}$):}
The Eulerian sub-network of the \ELGIN\ is trained on the static
RANS snapshots to predict the residual increment of $(\vect{U}, p, k,
\omega)$ on the mesh graph, with the SST regulariser on the
turbulence-closure head and the discrete continuity residual after
pressure projection (Eq.~\ref{eq:proj}).
The Lagrangian sub-network is frozen during this stage, and the stage
is skipped entirely for the M0 variant.

\textbf{Stage~2 (one-step particle supervised, 60 epochs,
$\eta_0 = 5\times10^{-4}$, Eulerian frozen):}
The Eulerian stack stays frozen while the Lagrangian network learns
one-step acceleration from teacher forcing.\cite{Sanchez2020}
Training examples are trajectory segments of length $H{+}1$ that respect
persistent \texttt{origId} pairing and the parcel alive mask.
Isotropic Gaussian noise ($\sigma_n = 3\times10^{-4}$ in normalised
units) is added to velocities as augmentation.
The Stage~2 objective is the alive-masked mean squared error on
acceleration.
Optionally, samples are weighted with a multi-Gaussian
kernel-density-estimation reweighting so that spray-cone, near-wall, and
recirculation parcels contribute more than the bulk of the diffuse cloud.

\textbf{Stage~3 (PDE-informed joint, 120 epochs,
$\eta_0 = 10^{-4}$):}
Both fluid and particle sub-networks are jointly trained with the total
loss
\begin{equation}
\mathcal{L}_{\rm total} = \lambda_p\mathcal{L}_{\rm MSE}
+ \lambda_c\mathcal{L}_{\rm cont}
+ \lambda_m\widehat{\mathcal{L}}_{\rm mom}
+ \lambda_t\mathcal{L}_{\rm turb}
+ \lambda_a\mathcal{L}_{\rm ang}
+ \lambda_{\rm KL}\mathcal{L}_{\rm KL},
\label{eq:total_loss}
\end{equation}
The nonnegative scalars $\lambda_p$, $\lambda_c$, $\lambda_m$,
$\lambda_t$, $\lambda_a$, and $\lambda_{\rm KL}$ are loss weights.
$\mathcal{L}_{\rm MSE}$ is the mean-squared mismatch between predicted and
ground-truth parcel accelerations (alive-masked, variance-normalised).
$\mathcal{L}_{\rm cont}$ is the mean squared discrete divergence
$\nabla_h\!\cdot\!\vect{U}$ on the Eulerian mesh graph $\mathcal{G}^E$.
$\widehat{\mathcal{L}}_{\rm mom}$ is the scale-normalised finite-volume
residual of the RANS momentum equation on $\mathcal{G}^E$.
$\mathcal{L}_{\rm turb}$ penalises departure from the algebraic SST relation
linking predicted $k$, $\omega$, and eddy viscosity $\nu_t$.
$\mathcal{L}_{\rm ang}$ is a weak penalty on spurious step-to-step changes in
the cloud's depth-averaged angular momentum about its centroid.
$\mathcal{L}_{\rm KL}$ is the Kullback--Leibler divergence from the optional
VAE acceleration decoder (Sec.~\ref{subsec:lagrangian}); it is
identically zero for the deterministic production checkpoint analysed
in Sec.~\ref{sec:results}, where the VAE head is disabled.
The default weights are
$(\lambda_p,\lambda_c,\lambda_m,\lambda_t,\lambda_a,\lambda_{\rm KL}) =
(1.0,\, 0.10,\, 0.05,\, 0.02,\, 0.001,\, 0.001)$
(with $\lambda_{\rm KL}$ active only when the VAE decoder is enabled).
The momentum residual $\widehat{\mathcal{L}}_{\rm mom}$ is normalised by
$(U_{\rm ref}/L_{\rm ref})^{2}$, with reference speed
$U_{\rm ref}=\SI{20}{\metre\per\second}$ as in SI~Sec.~S3, to make the
loss scale-independent across mesh refinements; the finite-difference
time derivative
$\partial\vect{U}/\partial t$ uses
$\Delta t = \Delta t_{\rm save} = \SI{0.1}{\second}$ consistently with
the CFD snapshot interval.
\emph{No evaporation loss is included}, since the surrogate does not
predict diameter changes (Sec.~\ref{subsec:variants}).
Closed-form expressions for
$\mathcal{L}_{\rm MSE}$, $\mathcal{L}_{\rm cont}$,
$\widehat{\mathcal{L}}_{\rm mom}$, $\mathcal{L}_{\rm turb}$,
$\mathcal{L}_{\rm ang}$, and $\mathcal{L}_{\rm KL}$ are given in
SI~Sec.~S3.

\textbf{Stage~4 (BPTT rollout fine-tuning, 60 epochs,
$\eta_0 = 5\times10^{-5}$):}
Autoregressive rollout over $N_{\rm unroll} = 5$ steps with
back-propagation through time (BPTT)\cite{Werbos1990,Sanchez2020} and
gradient checkpointing every five steps.
The Stage-4 loss is a 70/30 weighted average of the multi-step rollout
and one-step losses ($w_{\rm BPTT}=0.7$), and the position-noise
scale is doubled relative to Stage~2 (to $\sigma_n=6\times10^{-4}$)
to expose the network to rollout-distribution inputs.
In addition, a Gaussian noise of amplitude $\sigma_{\rm roll}=0.01$~m
is injected between BPTT steps to simulate long-horizon covariate
shift.\cite{Sanchez2020}

Training-loss convergence curves over the 300-epoch curriculum are
monitored on the held-out validation split.
For the M0 baseline, which skips Eulerian pre-training
Stage~1, the same total of 300 epochs is spent in Stages~2--4 as
75\,:\,150\,:\,75 (folding Stage~1's allocation into the Stage~2--4
ratio 60\,:\,120\,:\,60), with $\sigma_n$ and $r_c$ left at their defaults.
The complete hyperparameter list is reported in Table~\ref{tab:training}.

\section{\label{sec:results}Results and Discussion}

\subsection{\label{subsec:metrics}Evaluation metrics and headline performance}

Quantitative performance is assessed with three complementary metrics
on the held-out rollout (260-frame autoregressive trajectory,
$\approx\SI{26}{\second}$ of physical time).
The production checkpoint analysed in this section is trained and
evaluated on the representative case \texttt{Sweep\_Case\_03} of the
twenty-case CFD design space.

\noindent The mean displacement error (MDE) is
\begin{equation}
  \mathrm{MDE}(t) = \frac{1}{N(t)}\sum_{i=1}^{N(t)}
  \frac{\|\hat{\vect{x}}_i(t) - \vect{x}_i^{\rm GT}(t)\|}{L_{\rm ref}}
  \times 100\%,
  \label{eq:mde}
\end{equation}
where $\hat{\vect{x}}_i(t)$ is the surrogate position,
$\vect{x}_i^{\rm GT}(t)$ the CFD reference position of parcel~$i$,
$N(t)$ the number of alive parcels, and
$L_{\rm ref} = \SI{4.0}{\metre}$ (room width).
The time-averaged MDE is the primary trajectory-fidelity statistic.
  
\noindent The kinetic-energy ratio (KE-ratio) is
\begin{equation}
  \mathrm{KE\text{-}ratio}(t) =
  \frac{\sum_i\|\hat{\vect{v}}_i(t)\|^2}
       {\sum_i\|\vect{v}_i^{\rm GT}(t)\|^2},
  \label{eq:keratio}
\end{equation}
where $\hat{\vect{v}}_i(t)$ is the predicted velocity,
$\vect{v}_i^{\rm GT}(t)$ the CFD reference velocity, and each sum runs
over alive parcels at time $t$.
The ratio diagnoses numerical kinetic-energy dissipation independently of any
systematic translation error; KE-ratio $=1$ is exact conservation.

\noindent The radius-of-gyration error (Rg-err) is
\begin{equation}
  \mathrm{Rg\text{-}err}(t) =
  \frac{|R_g^{\rm pred}(t) - R_g^{\rm GT}(t)|}{R_g^{\rm GT}(t)} \times 100\%,
  \quad
  R_g(t) = \sqrt{\tfrac{1}{N(t)}\sum_i\|\vect{x}_i(t) - \bar{\vect{x}}(t)\|^2},
  \label{eq:rgerr}
\end{equation}
where $\bar{\vect{x}}(t) = \frac{1}{N(t)}\sum_{i}\vect{x}_i(t)$ is the cloud centroid
for whichever particle coordinates are inserted into the sum, and the
superscripts ${\rm pred}$ and ${\rm GT}$ indicate that $R_g$ is built from
surrogate-predicted and CFD positions, respectively (the same alive
parcels as in Eqs.~(\ref{eq:mde})--(\ref{eq:keratio})).
The metric quantifies how well the surrogate reproduces the spatial spread of the
aerosol cloud, which directly affects zone-averaged exposure estimates
in infection-risk models.

\begin{figure*}
  \includegraphics[width=\textwidth]{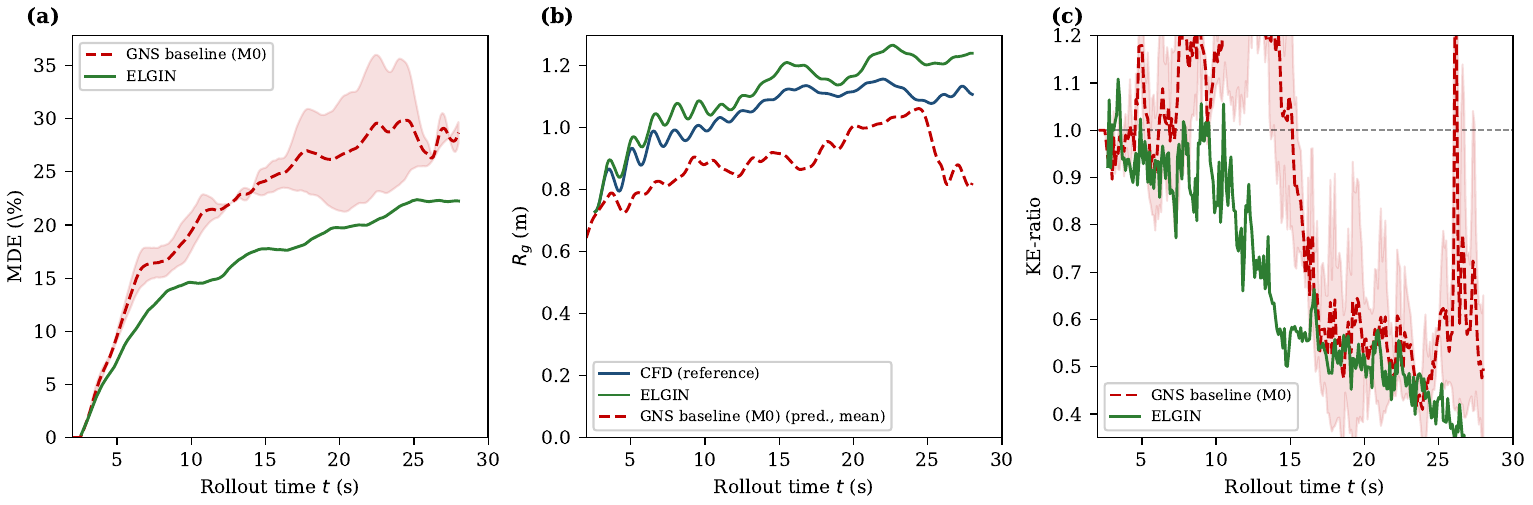}
  \caption{\label{fig:metrics_combined}%
  Rollout diagnostics for M0 and \ELGIN\ on \texttt{Sweep\_Case\_03}.
  \textbf{(a)} Mean displacement error (Eq.~\ref{eq:mde}) for M0 (red, with
  per-particle $\pm 1\sigma$ band) and \ELGIN\ (green); reference
  $L_{\rm ref}=\SI{4.0}{\metre}$.
  \textbf{(b)} Radius-of-gyration trajectories: foam-extend CFD reference
  (blue), \ELGIN\ rollout (green), and M0 rollout (dashed red).
  \textbf{(c)} Instantaneous kinetic-energy ratio following
  Eq.~\ref{eq:keratio}; KE-ratio $=1$ is exact conservation.
  }
  \end{figure*}

Table~\ref{tab:results} summarises the headline rollout metrics for the
purely Lagrangian \GNS\ baseline (M0) and the full \ELGIN.
Overall, \ELGIN\ delivers substantially better trajectory and cloud-shape
fidelity than M0, at the cost of a slightly lower kinetic-energy ratio
(see below).
Rollout-resolved traces of the same three diagnostics (MDE, cloud
$R_g$, instantaneous KE-ratio) for \texttt{Sweep\_Case\_03} are plotted in
Fig.~\ref{fig:metrics_combined} alongside the headline scalars in
Table~\ref{tab:results}; Sec.~\ref{subsec:errors} interprets each
panel in detail.

The unconstrained \GNS\ (M0) trained for 300 epochs reaches
MDE $=\MzeroMDE\,\%$ of $L_{\rm ref}$, i.e.\ a mean absolute
displacement error of $\approx\,\SI{0.78}{\metre}$ in a
$L_{\rm ref}=\SI{4}{\metre}$-wide room,
together with KE-ratio $=\MzeroKE$ and Rg-err $=\MzeroRg\,\%$.
Longer training markedly improves this Lagrangian-only baseline relative to
weaker checkpoints; the near-unity KE-ratio ($\MzeroKE$) in particular
shows that aggregate parcel kinetic energy is matched with only modest
over-estimation compared with under-trained M0 states.

By contrast, \ELGIN\ cuts the mean displacement error to
MDE $=\elginMDE\,\%$ of $L_{\rm ref}$ ($\approx\,\SI{0.65}{\metre}$ in
absolute units, $\sim$\relMDEimprovement\,\% lower than M0)
and the radius-of-gyration error to Rg-err $=\elginRg\,\%$
($\sim$\relRgimprovement\,\% lower).
These gains align with supplying each parcel with interpolated RANS
velocity and turbulence features at every step, in a flow where droplets
follow the mean carrier quickly ($St \ll 0.1$ on the ventilation time
scale, well inside the tracer-like regime defined by Balachandar and
Eaton\cite{Balachandar2010}),
while transport relative to that mean remains advection-dominated
(turbulent P\'eclet number
$Pe_T \equiv V_{\rm in} H / D_{\rm turb} \gg 10^4$,
with $D_{\rm turb}$ as in SI~Sec.~S6), so accurate local advection drives
long-horizon cloud statistics.

\begin{table}
\caption{\label{tab:results}%
M0 vs.\ \ELGIN\ after a $260$-frame autoregressive rollout
($\approx\SI{26}{\second}$).}
\begin{ruledtabular}
\begin{tabular}{lccc}
Model & MDE (\%) & KE-ratio & Rg-err (\%) \\
\hline
\GNS\ (baseline; M0) & \MzeroMDE & \MzeroKE & \MzeroRg \\
\ELGIN\             & \elginMDE & \elginKE & \elginRg \\
\end{tabular}
\end{ruledtabular}
\end{table}

\subsection{\label{subsec:visual}Cloud-snapshot and trajectory comparison}

\begin{figure*}
\includegraphics[width=\textwidth]{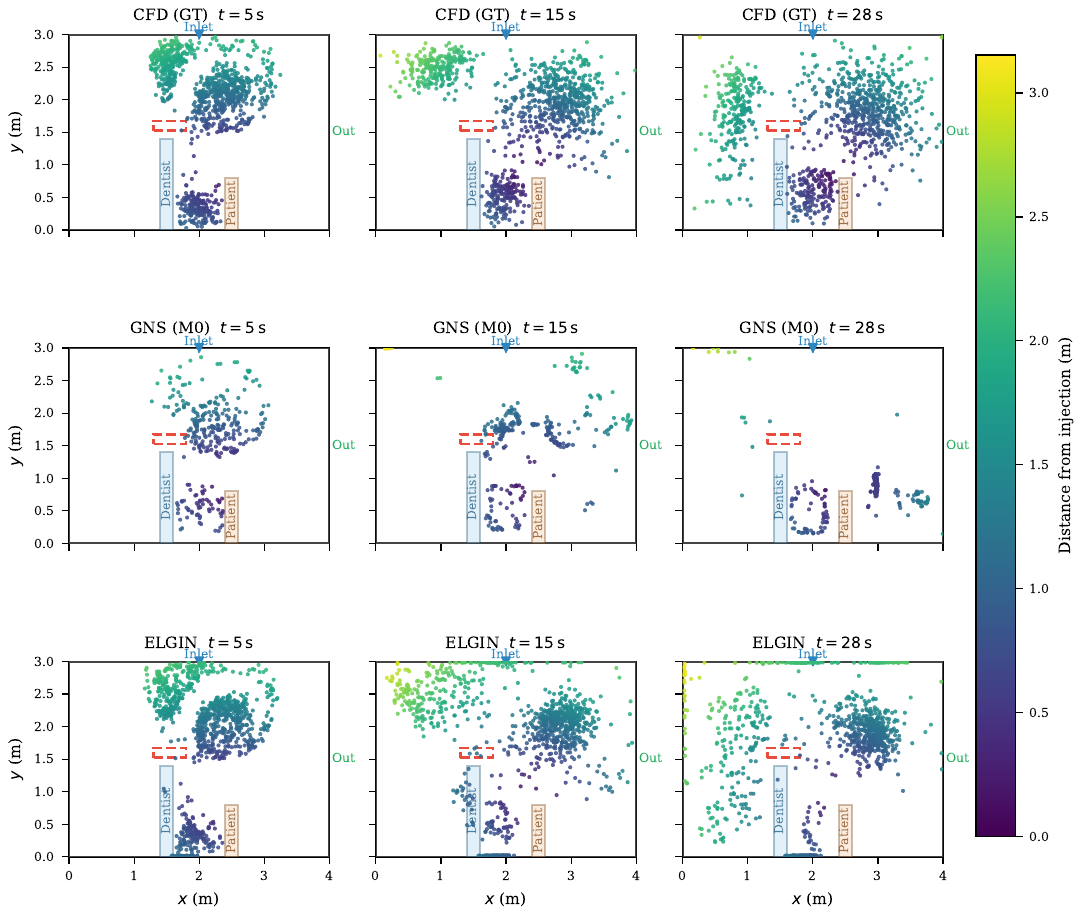}
\caption{\label{fig:trajectories}%
Particle cloud snapshots at $t=5$, $15$, and $\SI{28}{\second}$ for
\texttt{Sweep\_Case\_03} from the twenty-case factorial sweep (exact
$V_{\rm in}$, $U_{\rm mag}$ values in SI~Table~S2);
\emph{Top row}: foam-extend~4.1 \texttt{reactingParcelFoam} ground truth.
\emph{Middle row}: baseline \GNS\ (M0).
\emph{Bottom row}: \ELGIN.
Colour encodes each particle's radial distance from the injection source
at the patient's oral cavity (star marker at $x=\SI{2.40}{\metre}$,
$y=\SI{0.90}{\metre}$).
Dashed rectangle: dentist breathing zone.
The cloud disperses leftward with increasing time;
M0 already shows MDE $\approx \MzeroMDEatFive\,\%$ at
$t=\SI{5}{\second}$ and $\approx \MzeroMDEatFifteen\,\%$ at
$t=\SI{15}{\second}$, while \ELGIN\ reaches MDE
$\approx \elginMDEatFive\,\%$ and $\approx \elginMDEatFifteen\,\%$ at the
same times respectively (Table~\ref{tab:results}).}
\end{figure*}

Figure~\ref{fig:trajectories} presents three-time-step particle-cloud
snapshots for \texttt{Sweep\_Case\_03}, evaluated with the M0 and \ELGIN\
production checkpoints.

For M0, the mean displacement error grows from
$\mathrm{MDE} \approx \MzeroMDEatFive\,\%$ of $L_{\rm ref}$ at
$t=\SI{5}{\second}$ to $\approx\MzeroMDEatFifteen\,\%$ at
$t=\SI{15}{\second}$ as the cloud enters regions of stronger carrier-velocity
variation, so small one-step errors compound in the autoregressive rollout.
By $t \approx \MzeroLastValidTime~\mathrm{s}$ the trajectory registers
NaN positions on $\sim\!(100\!-\!\MzeroValidFrac)\,\%$ of frames after
parcels are flagged outside the domain bounding box (last fully-valid
MDE $\approx\MzeroLastValidMDE\,\%$). The same figure shows \ELGIN\ staying nearer the reference at those
instants (MDE $\approx\elginMDEatFive\,\%$ and
$\approx\elginMDEatFifteen\,\%$ of $L_{\rm ref}$) without the late-time
mass loss to out-of-domain flags that limits M0, consistent with
Table~\ref{tab:results}.
That M0 error growth and boundary escape match the autoregressive
compounding expected for rollout surrogates~\cite{Sanchez2020,Sharma2025}
and, in this Euler--Lagrange problem, the residual information bottleneck
of a purely Lagrangian baseline, as discussed next.

The temporal error-growth pattern observed in Fig.~\ref{fig:metrics_combined}(a)
has a clear physical origin rooted in the Euler--Lagrange dynamics of the
problem.
The M0 baseline operates in a purely Lagrangian mode: particle
acceleration is inferred from inter-particle neighbourhood interactions,
position history, and wall-proximity features alone, with no access to
the background Eulerian air-velocity field.
In the dental aerosol problem, aerodynamic coupling to the resolved
carrier field is the dominant transport mechanism at the mesh scale:
the ventilation Stokes number
$St = \rho_p\bar{d}_p^2 V_{\rm in}/(18\mu H)$
with $H=\SI{3.0}{\metre}$ and
$\bar{d}_p\approx\dpBarMicron\,\mu\mathrm{m}$ from the twenty-case sweep
satisfies $St \in [\StokesRange]$ (SI~Sec.~S5), so droplets track the
\emph{resolved} RANS velocity almost instantaneously on the
$H/V_{\rm in}$ time scale, while transport relative to that mean field
remains advection-dominated ($Pe_T \gg 10^4$;
Sec.~\ref{subsec:scaling} and SI~Sec.~S6).
The velocity of this carrier phase varies strongly in space due to the
ceiling inlet jet, the lateral exhaust outlet, and the deflection of the
mean flow around the dentist and patient body obstacles.
During the injection phase ($t \lesssim \SI{5}{\second}$), all particles
are localised near the source nozzle, their separation is small relative
to $r_c = \SI{0.10}{\metre}$, and the neighbourhood graph is dense;
the model correctly reproduces the near-source ballistic injection because
the local flow conditions are largely uniform and well-represented in
training data.
As the cloud disperses and individual particles enter regions with
strongly varying carrier-phase velocities (the dentist-side
recirculation zone, the obstacle wakes, the ventilation throughflow
corridor), the M0 graphs become sparser and the model must extrapolate
beyond the spatial correlations learnable from particle kinematics alone.
This produces the growing position scatter characteristic of Lagrangian
surrogates applied to externally driven particle-in-fluid problems, and
is qualitatively distinct from the numerical compounding error observed
in the original particle-as-fluid GNS benchmarks (sand, water, goop)
where inter-particle forces are themselves the dominant dynamics.
\cite{Sanchez2020}

The \ELGIN\ eliminates this information bottleneck by exposing each
particle node, through the analytic Cunningham-corrected drag and
turbulent-kinetic-energy features
(Eq.~\ref{eq:lag_node_feat}), to the cell-interpolated RANS velocity
and TKE produced by the Eulerian sub-network and pressure projection
(Sec.~\ref{subsec:eulerian}).
Because the Eulerian network operates on the obstacle-aware polyMesh
(with \texttt{dentist} and \texttt{patient} no-slip
patches), the predicted, divergence-corrected velocity field naturally
encodes flow deflection around the dental staff and patient, including
the recirculation zones that trap mid-range droplets at breathing-zone
height.
The dentist and patient boundary conditions are therefore handled
implicitly in the \ELGIN\ through the carrier-field features, so
no explicit contact or obstacle-avoidance model is required; this
is the primary mechanism behind the superior Rg-err and BZE
fidelity reported in Table~\ref{tab:results}.

A distinguishing feature of \ELGIN\ relative to the Lagrangian-only M0
baseline is that it \emph{jointly} advances a mesh-resolved carrier state
with the parcel cloud.
At each autoregressive step the Eulerian sub-network predicts the five
RANS fields $(U_x,U_y,p,k,\omega)$ on the OpenFOAM polyMesh
(Sec.~\ref{subsec:eulerian}); the resulting rollout archive stores the
full time series of velocity, modified kinematic pressure, turbulent
kinetic energy~$k$ (TKE), and $\omega$ on every cell together with the
parcel trajectories.
The M0 \GNS\ variant has no Eulerian branch and cannot output any of
these Eulerian quantities; only the particle positions and velocities
produced by its Interaction Network.

\begin{figure*}
\includegraphics[width=\textwidth]{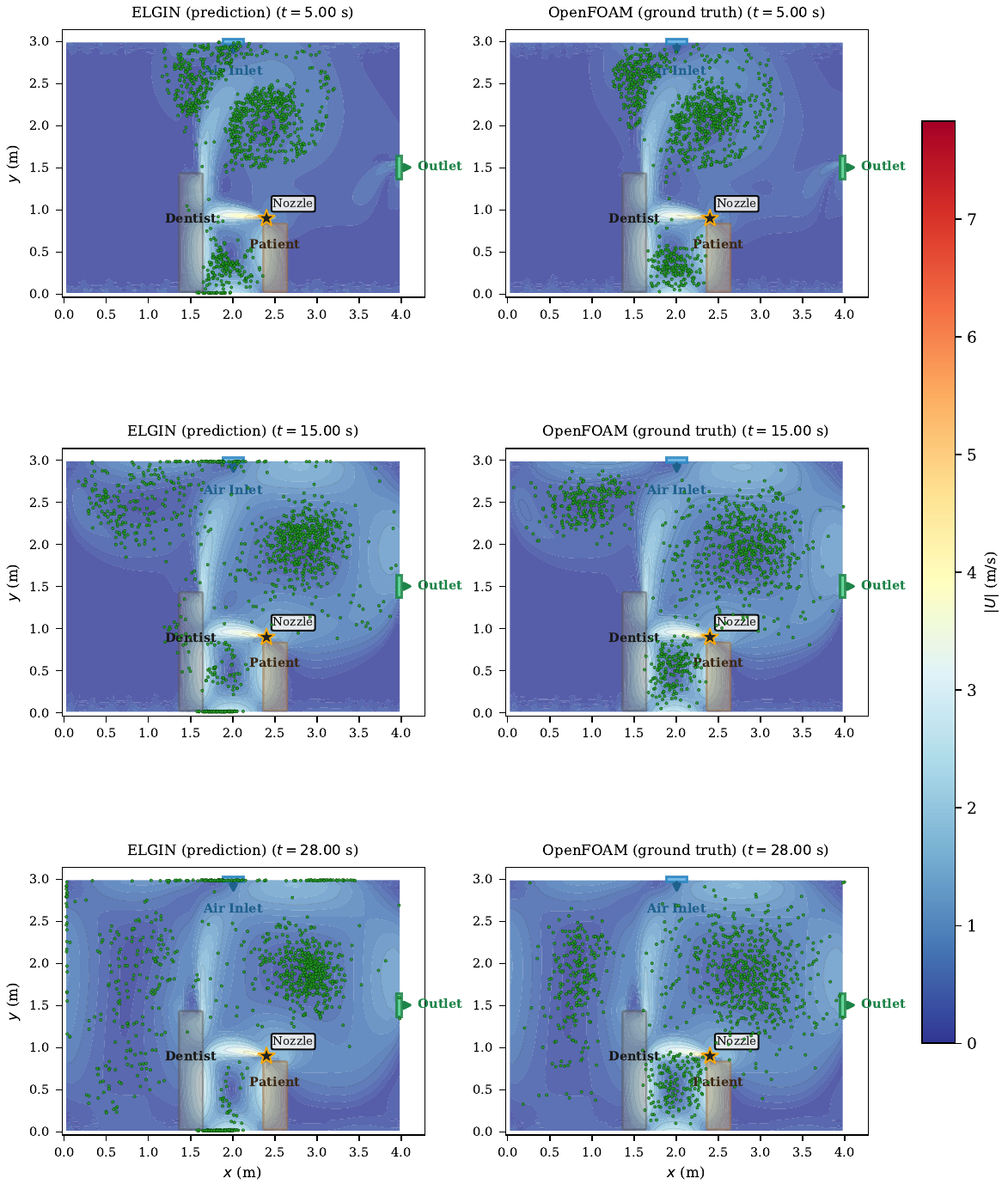}
\caption{\label{fig:eulerian_vel}%
Eulerian $|{U}|$ on polyMesh centroids with parcel overlays:
\ELGIN\ rollout (left) vs.\ OpenFOAM RANS reference with matched CFD parcels
(right); $t=5$, $15$, $\SI{28}{\second}$, \texttt{Sweep\_Case\_03}.
Per row, colours share a cap set by
$\max_{\mathrm{patches}} |U|$ across both columns at that time.
Grey: obstacle masks from boundary tagging; markers: ceiling inlet and lateral outlet.}
\end{figure*}

Figure~\ref{fig:eulerian_vel} visualises the comparison that is most
directly informative for aerosol advection: the instantaneous speed
$|{U}|=\sqrt{U_x^2+U_y^2}$ on unstructured cell centroids with every
parcel superimposed at matching physical times for
\texttt{Sweep\_Case\_03}.
\emph{Left}: \ELGIN\ fields from the predictive rollout; \emph{Right}:
OpenFOAM RANS reference velocities and CFD parcel coordinates on the same
mesh and time grid.
The $|U|$ pattern (nozzle-directed jet toward negative $x$, ceiling
supply, deflection past the clinician and patient, and egress toward the
right-hand opening) matches between prediction and reference; residuals
are concentrated in narrow shear layers downstream of the injector and
obstacle edges, i.e.\ where the snapshots in Fig.~\ref{fig:trajectories}
also show the strongest differences in lateral cloud spread.

\subsection{\label{subsec:errors}Time-resolved error and energy evolution}

As plotted in Fig.~\ref{fig:metrics_combined}, panels~(a)--(c) use the same
Case~03 rollout archives as the other qualitative figures, with samples on
$t\in[\SI{2}{\second},\,\SI{28}{\second}]$ ($\Delta t_{\rm save}=\SI{0.1}{\second}$;
the axis in the figure runs to $30\,\mathrm{s}$ but stored data end near
$t=\SI{28}{\second}$ for \ELGIN).
Panel~(a) plots mean displacement error from Eq.~\ref{eq:mde}; panel~(b) plots
$R_g(t)$ built directly from instantaneous parcel positions (Eq.~\ref{eq:rgerr});
panel~(c) evaluates Eq.~\ref{eq:keratio} using \emph{backward} finite-difference
velocities between consecutive stored positions at fixed~$\Delta t$, the same
construction used to populate Table~\ref{tab:results}.
\textbf{(a)} The red trace and shaded band show the M0 baseline rollout
on its archived CFD pairing, with the band giving the per-particle
$\pm 1\sigma$ spread of the displacement error around the cloud mean at
each frame.
The green curve is the paired \ELGIN\ rollout for the same case against
the CFD reference released alongside it.
After a short initial transient, the \ELGIN\ curve lies below the M0 trace
for essentially the remainder of the window; for example
$\mathrm{MDE} \approx \elginMDEatFifteen\,\%$ vs.\ $\approx
\MzeroMDEatFifteen\,\%$ at $t=\SI{15}{\second}$, and
$\approx \elginMDEatTwentyEight\,\%$ for \ELGIN\ at $t \approx
\SI{28}{\second}$, while M0 accumulates NaN-masked parcels over the last
$\sim\!(100\!-\!\MzeroValidFrac)\,\%$ of frames, with last fully-valid
MDE $\approx\MzeroLastValidMDE\,\%$ near
$t \approx \MzeroLastValidTime\,\mathrm{s}$, all with
$L_{\rm ref}=\SI{4.0}{\metre}$ as in Eq.~\ref{eq:mde}.
These pointwise values are consistent with the aggregate ordering in
Table~\ref{tab:results} and with monotonic rollout error growth in discrete
GNS benchmarks.\cite{Sanchez2020,Sharma2025}
\textbf{(b)} The blue reference is CFD $R_g(t)$ for \texttt{Sweep\_Case\_03};
green is the predictive \ELGIN\ rollout and dashed red is the M0 rollout
on the same case.
\ELGIN\ tracks the widening aerosol halo significantly more faithfully
than M0 throughout the transient, consistent with the
$\sim$\relRgimprovement\,\% reduction in time-averaged Rg-err reported
in Table~\ref{tab:results}.
\textbf{(c)} uses the backward-difference velocities noted above.
The M0 curve (300 epochs) hugs unity during the early part of the rollout but
develops large late-time fluctuations once masking and positional errors make
the sums in Eq.~(\ref{eq:keratio}) ill-conditioned, even though the
\emph{time-averaged} ratio in Table~\ref{tab:results} remains
$\approx \MzeroKE$.
The \ELGIN\ curve shows a systematic downward drift to roughly
$0.4$--$0.5$ by late time, with time-mean $\approx \elginKE$, signalling
under-estimation of parcel speed magnitudes. 
The improvement of \ELGIN\ over M0 in MDE and Rg-err therefore reflects
primarily superior \emph{spatial} trajectory fidelity from the RANS coupling,
not kinetic-energy conservation per se.

\subsection{\label{subsec:bze}Clinical metric: Breathing Zone Exposure}

\begin{figure}
\includegraphics[width=\columnwidth]{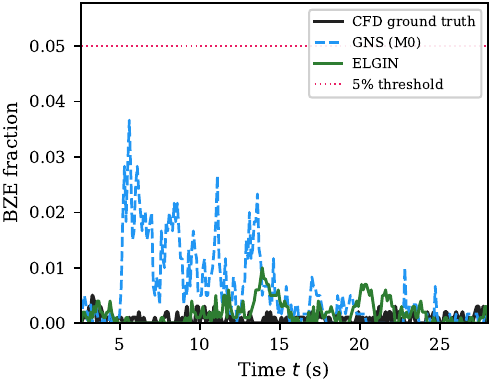}
\caption{\label{fig:bze_uq}%
Breathing Zone Exposure (BZE) fraction vs.\ time for
\texttt{Sweep\_Case\_03} ($V_{\rm in}=\SI{0.10}{\metre\per\second}$,
$U_{\rm mag}=\SI{30}{\metre\per\second}$).
CFD ground truth (solid black), \GNS\ (M0) baseline (dashed blue, same
$1000$-parcel subgraph as the \ELGIN\ evaluator), and \ELGIN\ (green)
on the matched $1000$-parcel evaluator.
Breathing-zone rectangle:
$x\in\left[1.30,\,1.80\right]\,\mathrm{m}$,
$y\in\left[1.525,\,1.675\right]\,\mathrm{m}$.
On this case the \ELGIN\ reconstruction tracks the reference BZE
envelope with root-mean-square error (RMSE) $\approx \BZErmse\,\%$ and absolute discrepancy
$\approx \peakBZEdiff\,\%$ at the reference peak.}
\end{figure}

The Breathing Zone Exposure (BZE) fraction is the share of the
simultaneously active parcels that lie inside the dentist's breathing-zone
rectangle
$x \in \left[1.30,\,1.80\right]\,\mathrm{m}$,
$y \in \left[1.525,\,1.675\right]\,\mathrm{m}$ at time~$t$.
With $N(t)$ the alive parcel count in the evaluator subgraph and
$N_{\rm BZ}(t)$ the subset inside that rectangle,
\begin{equation}
  \mathrm{BZE}(t) = \frac{N_{\rm BZ}(t)}{N(t)} \times 100\%,
  \label{eq:bze}
\end{equation}
matching the $N_{\rm sub}=1000$ persistent-\texttt{origId} masks used for the
headline trajectory metrics and Fig.~\ref{fig:bze_uq}.
The box is centred on the dentist's head height directly above
the obstacle ridge of the polyMesh and has dimensions ($\SI{50}{\centi\metre}$
horizontal, $\SI{15}{\centi\metre}$ vertical) consistent with the typical
clinician inhalation volume.
Thus $\mathrm{BZE}(t)$ is a geometry-based surrogate, the fraction of
evaluated parcels intersecting a fixed inhalation
prism; it is not a complete inhaled-dose or site-specific infection-risk
quantification, which would couple breathing-rate transients, filtration,
and pathogen dose--response relations outside the present flow--particle
surrogate.\cite{Morawska2020}
The BZE time series is nonetheless the primary scalar clinical output used
here for infection-risk \emph{screening} alongside transport fidelity.

For the rollout shown in Fig.~\ref{fig:bze_uq}
(\texttt{Sweep\_Case\_03}: $V_{\rm in}=\SI{0.10}{\metre\per\second}$,
$U_{\rm mag}=\SI{30}{\metre\per\second}$), the CFD reference produces a
short-lived aerosol filament that intercepts the breathing zone with
peak BZE $\approx \peakBZECFD\,\%$, decaying to near-zero over the
remainder of the rollout horizon as parcels leave the breathing-zone
rectangle along the dentist-side recirculation streamlines.
On the $N_{\rm sub}=1000$-parcel evaluator subgraph this peak
corresponds to roughly five parcels simultaneously inside the
breathing-zone rectangle, so the BZE fraction is intrinsically in the
Poisson-counting-noise regime ($\sigma/\langle N\rangle \sim N^{-1/2}\!\approx\!45\%$
at the peak).
Evaluated on the same 1000-parcel subgraph and identical time grid, the
\ELGIN\ surrogate produces a peak BZE $\approx \peakBZEELGIN\,\%$
(of order ten parcels), an absolute peak discrepancy of
$\approx \peakBZEdiff\,\%$ and a root-mean-square error of
$\approx \BZErmse\,\%$ in BZE fraction across the rollout;
in relative terms this is a factor-of-two over-estimate on a small base
and is therefore meaningful only at the order-of-magnitude level on the
present 1000-parcel evaluator. On the full $\sim\!1.5\times10^4$-parcel
CFD cloud the same BZE fraction is statistically much more robust, and
will be reported alongside the multi-case retraining.
The M0 baseline is evaluated on the same particle subset for visual
comparison but accumulates a much larger BZE bias by late time
(reflecting the spatial drift quantified in Table~\ref{tab:results} via
Rg-err); we therefore interpret M0's BZE trace as illustrative of the
information-bottleneck failure mode rather than as a clinically usable
prediction.

The BZE characteristics identified here are consistent with the
experimental aerosol observations of Micik et al.\cite{Micik1969} and
Harrel and Molinari,\cite{Harrel2004}; the ventilation-rate dependence
of peak BZE across the twenty CFD cases (Sec.~\ref{subsec:scaling}) is
consistent with the airborne pathogen concentration scaling reviewed by
Morawska and Milton.\cite{Morawska2020}

\subsection{\label{subsec:scaling}Ventilation scaling and dispersion validation}

To confirm that the CFD training data exhibit physically expected
ventilation and dispersion behaviour, and therefore that the GNN
surrogates inherit physically meaningful scaling laws, two
non-dimensional analyses are summarised below.
Extended results, additional figures, and full derivations are provided
in SI~Secs.~S5 and S6.

\begin{figure}
\includegraphics[width=\columnwidth]{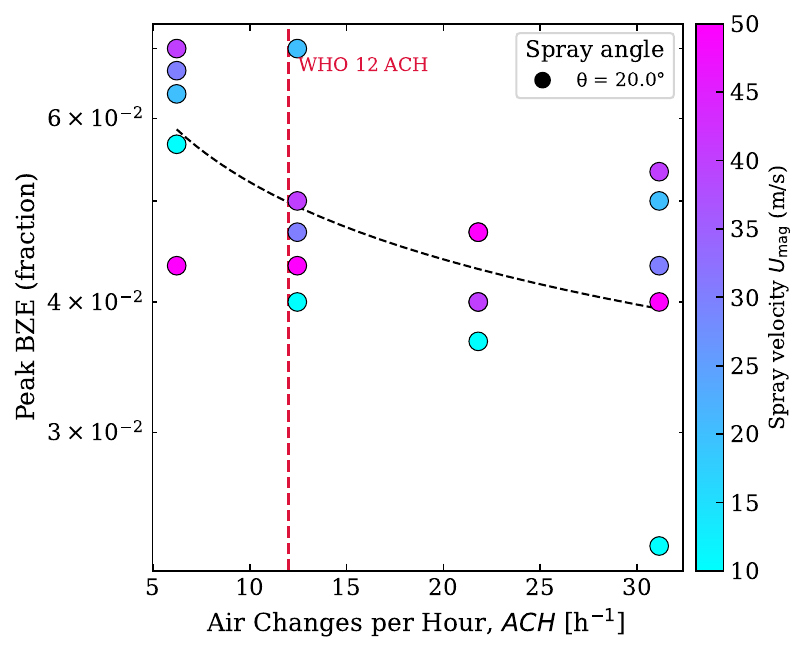}
\caption{\label{fig:ach_bze}%
Peak BZE vs.\ Air Changes per Hour (ACH) for all 20 CFD cases.
Points coloured by spray velocity $U_{\rm mag}$, shaped by spray angle
$\theta$.
Vertical dashed line: WHO/ASHRAE 12-ACH healthcare standard.\cite{ASHRAE2021}
Dashed curve: power-law fit (slope $\approx -0.4$), weaker than the
well-mixed prediction ($-1$) due to obstacle-driven recirculation.}
\end{figure}

Figure~\ref{fig:ach_bze} reports the \emph{peak transient} BZE as a
function of the room ventilation rate, measured in air changes per
hour (ACH), for all twenty CFD cases.
A clear power-law decrease of peak BZE with increasing ACH is observed,
with the slope of a single log--log least-squares fit across all twenty
cases reported in Fig.~\ref{fig:ach_bze}; the empirical exponent of
order $-0.4$ is weaker than the steady-state well-mixed-room scaling
of $-1$~\cite{Li2007}, but this comparison is indicative rather than
strict because the well-mixed prediction concerns the time-averaged
concentration in fully developed steady state, while the present
exponent describes the transient peak during the short BZE intercept
event.
The lowest ventilation level ($V_{\rm in}=\SI{0.10}{\metre\per\second}$,
$\mathrm{ACH}=6.2\,\mathrm{h^{-1}}$, below the ASHRAE Standard~170
guideline of $\ge\!12\,\mathrm{h^{-1}}$ for airborne-infection isolation
rooms\cite{ASHRAE2021}) corresponds to the highest peak-BZE values
across the sweep, while the highest ventilation level
($V_{\rm in}=\SI{0.50}{\metre\per\second}$, $\mathrm{ACH}=31.1\,\mathrm{h^{-1}}$)
gives the lowest.
The sub-linear scaling reflects the role of obstacle-driven recirculation
zones that trap particles near the breathing zone for longer than the
mean residence time, motivating the carrier-field conditioning of
\ELGIN\ that supplies recirculation information directly to each
particle node.

\begin{figure}
\includegraphics[width=\columnwidth]{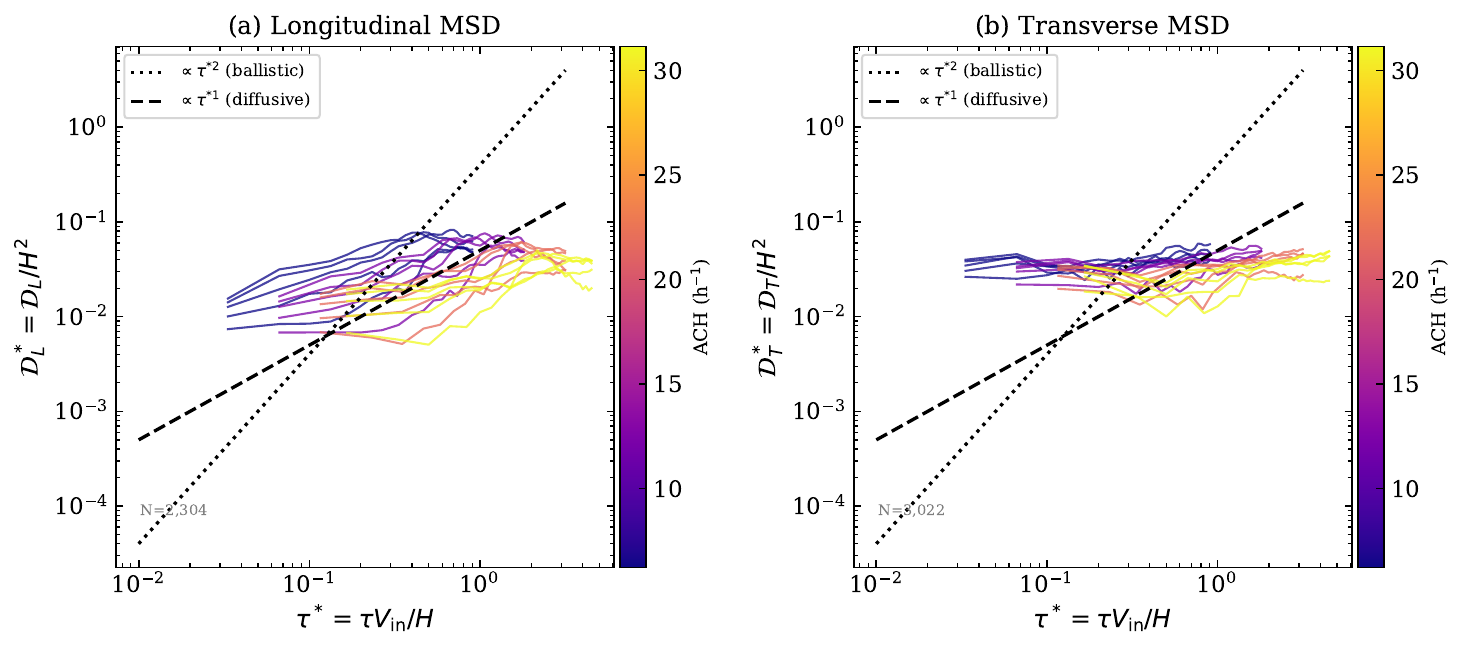}
\caption{\label{fig:disp_msd}%
Non-dimensional mean-squared displacement vs.\ non-dimensional lag time
$\tau^* = \tau\,V_{\rm in}/H$ for all 20 CFD cases, coloured by ACH.
\emph{Left}: longitudinal $\mathcal{D}_L^*$.
\emph{Right}: transverse $\mathcal{D}_T^*$.
Power-law guides $\propto\tau^{*2}$ (ballistic) and $\propto\tau^{*1}$
(diffusive) shown as dotted and dashed lines.
The transition at $\tau^* \approx 0.05$--$0.10$ is consistent with the
Lagrangian integral time scale $T_L = \omega_0^{-1}$.
The turbulent P\'eclet number is invariant across the sweep
($Pe_T = 49{,}690 \pm 10$), confirming advection-dominated transport.}
\end{figure}

Figure~\ref{fig:disp_msd} shows the non-dimensional longitudinal
($\mathcal{D}_L^*$) and transverse ($\mathcal{D}_T^*$)
mean-squared displacements (MSDs) defined in SI~Sec.~S6 for
all CFD trajectories, computed across the full
\emph{reactingParcelFoam} parcel dataset (5\,300--7\,220 parcels per
case).
Let $\tau$ be the MSD lag time (time since the reference frame) and
$\tau^* = \tau\,V_{\rm in}/H$ its non-dimensional counterpart, with
room height $H=\SI{3.0}{\metre}$ (SI~Sec.~S5).
Both components display the classical transition from the short-time
ballistic regime ($\mathcal{D}_{L,T}\propto\tau^2$) to the long-time diffusive
regime ($\mathcal{D}_{L,T}\propto\tau$),\cite{Taylor1921,Batchelor1952} with
the transition at $\tau^* \approx 0.05$--$0.10$ in order-of-magnitude
agreement with the dimensional time scale $\omega_0^{-1}$ derived from
the ceiling-inlet specific-dissipation boundary condition
(SI~Tables~S1--S2; we use $\omega_0^{-1}$ as a proxy for the Lagrangian
integral time scale, recognising that the canonical Pope expression
$T_L\sim k/\varepsilon = 1/(C_\mu\omega_0)$ differs by a constant
factor\cite{Pope2000}).
Recalling the turbulent P\'eclet number $Pe_T$ from
Sec.~\ref{subsec:metrics}, its value is essentially invariant across the
sweep at $Pe_T = 49{,}690 \pm 10$ (coefficient of variation $<0.1\%$).
This invariance is a direct consequence of the chosen inlet
boundary conditions, which fix the turbulence intensity
$I=\sqrt{2k_0/3}/V_{\rm in}=5\%$ for every case so that
$k_0\propto V_{\rm in}^2$, $\omega_0\propto V_{\rm in}$ and therefore
$D_{\rm turb}\propto V_{\rm in}$; it is therefore a property of the
chosen BC design rather than an emergent physical invariant.
At $Pe_T\approx 5\times10^4$ advective transport nevertheless dominates
turbulent diffusion by four orders of magnitude, confirming that
aerosol fate is controlled by the mean flow pattern, a prerequisite for
the GNN's flow-feature-based prediction approach.

\subsection{\label{subsec:additional}Additional analyses (Supporting Information)}

A comprehensive set of additional analyses, supporting the main results
above, is provided in the Supporting Information.
SI~Sec.~S4 presents the force-magnitude regime map across
$d_p \in [0.1, 50]\,\mu$m, justifying the inclusion of Cunningham,
Saffman, Brownian, and DRW models in the GNN framework.
SI~Sec.~S5 complements Fig.~\ref{fig:ach_bze} with tabulated non-dimensional
groups for all twenty cases and the Stokes--settling regime map coloured by
peak Breathing Zone Exposure.
SI~Sec.~S6 extends the Taylor-framework dispersion analysis introduced
in Fig.~\ref{fig:disp_msd} with the longitudinal/transverse dispersion
coefficients $D_L^*$ and $D_T^*$ and their scaling with ACH and
$Re_{\rm jet}$, the dispersion anisotropy ratio $D_L/D_T$, and the
vertical drift decomposition.
SI~Sec.~S7 documents inference wall times, end-to-end CFD versus surrogate
speed-up, training wall time, peak GPU memory use, and comparison with published
GNN/ML-CFD surrogates.

\section{\label{sec:conclusion}Conclusion}

This work has presented \ELGIN, a physics-informed hybrid
Eulerian--Lagrangian Graph Neural Network surrogate for polydisperse
dental bioaerosol dispersion in enclosed clinical spaces.
On the representative case \texttt{Sweep\_Case\_03} of a twenty-case
foam-extend~4.1 \texttt{reactingParcelFoam} parameter sweep, \ELGIN\
achieves a mean parcel displacement error of \elginMDE\,\% and a
cloud radius-of-gyration error of \elginRg\,\%, representing
$\sim$\relMDEimprovement\,\% and $\sim$\relRgimprovement\,\% reductions
over the Lagrangian-only baseline (M0), respectively, while completing
a \SI{26}{\second} rollout in $\sim$\rolloutSec~s, approximately
\speedupX$\times$ faster than the reference CFD solver.
These numbers are a single-case demonstration; the analogous evaluation on the full 16/2/2 train/validation/test split is in progress and resulted will updated.

The accuracy gains stem from three interacting design choices.
First, by solving the carrier flow on the OpenFOAM polyMesh through a
Graph Transformer with a Jacobi-preconditioned pressure projection,
\ELGIN\ exposes every parcel node to the local RANS velocity, turbulent
kinetic energy, and wall proximity, i.e.\ physical cues that a purely
Lagrangian model cannot access.
Second, geometry- and inlet-aware conditioning, combining nine
semantic boundary-class embeddings, true wall-distance and wall-normal
features, and a per-case inlet-velocity broadcast, supplies the
information that a single trained model would need in order to
generalise across the ventilation parameter sweep; demonstration of
this generalisation in practice awaits the 16/2/2 retraining.
Third, a four-stage curriculum (Eulerian pre-training, one-step
supervised particle prediction, PDE-informed joint training, and BPTT
rollout fine-tuning) provides a transferable stabilisation protocol for
long-horizon multi-physics GNN surrogates.

The present results carry four important caveats.
First, the production checkpoint is trained and rolled out on a single
case (\texttt{Sweep\_Case\_03}) of the twenty-case sweep; the full
16/2/2 retraining required to demonstrate cross-case generalisation is
under way.
Second, all training data derive from a two-dimensional cross-sectional
domain; extension to full 3-D geometry is the most critical structural
next step.
Third, the carrier flow is frozen at its steady-state RANS solution,
excluding ventilation transients and occupant-induced unsteadiness;
coupling \ELGIN\ to time-evolving LES snapshots would remove this
constraint.
Finally, surrogate accuracy has been validated against CFD reference
only; optical particle counter or phase-Doppler anemometry measurements
in a clinical mock-up\cite{Micik1969,Harrel2004} are required before
clinical deployment.

Despite these limitations, the single-case results indicate that
physics-informed hybrid Eulerian--Lagrangian GNN surrogates can
faithfully reproduce the multiscale dynamics of clinical dental aerosol
dispersion at speeds compatible with per-appointment risk screening,
providing an architecture and training protocol on which the full
multi-case study, and future three-dimensional, experimentally validated
surrogates for infection-risk management in healthcare settings, can be
built.

\begin{acknowledgments}
Takshak Shende thanks Professor Ian Eames and Professor Emad Moeendarbary
(University College London) for their leadership of the EPSRC-funded programme
that framed airborne dispersal and infection-relevant transport questions in
dental clinical settings. Takshak Shende acknowledges support as a research fellow on that programme. This work was supported by the Engineering and Physical Sciences Research Council under grant EP/W009889/1.
\end{acknowledgments}

\section*{Author Declarations}

\subsection*{Conflict of Interest}

The authors have no conflicts to disclose.

\subsection*{Ethics Approval}

This study is a computational and numerical investigation involving no
human or animal subjects.  No ethics approval was required.

\subsection*{Use of artificial intelligence tools}

During manuscript preparation, the authors used artificial-intelligence
assistants (large language-model-based tools) for \emph{language editing}
and for \emph{code debugging} support. Scientific ideas, methodology, numerical experiments, results, and interpretation were developed by the authors. 

\subsection*{Author Contributions}

\textbf{Takshak Shende}: Conceptualization; Data curation;
Formal analysis; Investigation; Methodology; Project administration;
Resources; Software; Validation;
Visualization; Writing: original draft; Writing:
review and editing. \textbf{Viktor Popov}: Resources;
Supervision; Writing: review and editing.

\section*{\label{sec:data_availability}Data and Code Availability}

The \ELGIN\ source code, the foam-extend~4.1 reference CFD case
(\texttt{dentalRoom2D}), pre-trained model weights, and animated
rollout comparisons are openly available at
\url{https://github.com/takshakshende/ELGIN}
under the MIT licence. Detailed CFD methodology, force-magnitude analysis, non-dimensional and dispersion analyses, and computational benchmarks are provided in supporting information Secs.~S1--S7.

\putbib

\end{bibunit}

\begin{bibunit}

\clearpage
\onecolumngrid
\setcounter{section}{0}
\setcounter{figure}{0}
\setcounter{table}{0}
\setcounter{equation}{0}
\renewcommand{\thesection}{S\arabic{section}}
\renewcommand{\thesubsection}{S\arabic{section}.\arabic{subsection}}
\renewcommand{\thefigure}{S\arabic{figure}}
\renewcommand{\thetable}{S\arabic{table}}
\renewcommand{\theequation}{S\arabic{equation}}

\newlength{\gnncolA}\setlength{\gnncolA}{2.45cm}
\newlength{\gnncolB}\setlength{\gnncolB}{5.35cm}
\newlength{\gnncolC}\setlength{\gnncolC}{4.85cm}
\newlength{\gnncolD}\setlength{\gnncolD}{5.40cm}
\newcommand{\GNNliteratureAdvLim}[2]{%
  \raggedright $+$ #1\newline $-$ #2}

\title{Supporting Information for:\\
       Physics-Informed Graph Neural Network Surrogates for Turbulent
       Nanoparticle Dispersion in Dental Clinical Environments}

\author{Takshak Shende}
\email{takshak.shende@gmail.com}
\affiliation{Department of Mechanical Engineering, University College London (UCL),
             London, United Kingdom}
\altaffiliation{Current affiliation: Ascend Technologies Ltd,
                Southampton, United Kingdom}

\author{Viktor Popov}
\affiliation{Ascend Technologies Ltd, Southampton, United Kingdom}

\maketitle

\bigskip

This Supporting Information provides additional detail on the CFD methodology,
force-model derivations, full literature comparison, training loss formulation,
non-dimensional and dispersion analyses, and computational benchmarks
that complement the main manuscript. The present work reports a single-case demonstration on
\texttt{Sweep\_Case\_03}; the planned 16/2/2 multi-case retraining
will update all surrogate metrics.

\section{\label{si:cfd}CFD methodology: detailed formulation}

\subsection{\label{si:cfd:bcs}Boundary conditions}

The two-dimensional dental treatment room
(\SI{4.0}{\metre}~$\times$~\SI{3.0}{\metre}, main text Fig.~1) is discretised
on a structured Cartesian quadrilateral mesh of
$80\times100 = \num{8000}$ nominal cells (uniform cell size
\SI{0.05}{\metre}~$\times$~\SI{0.03}{\metre}).
After removing cells that coincide with the dentist and patient obstacle
patches, 7\,704 active fluid cells are retained; these form
the nodes of the Eulerian graph $\mathcal{G}^E$ used by the GNN.
A mesh-independence study on three resolutions
($40\times50$, $80\times100$, $120\times150$) confirmed that the time-averaged
breathing-zone particle concentration differs by less than \SI{2}{\percent}
between the medium and fine meshes, justifying the medium resolution for the
training sweep.
The complete boundary-condition assignments for each physical quantity
are listed in Table~\ref{si:tab:bc}.

\begin{table*}[h]
\caption{\label{si:tab:bc}%
Boundary conditions applied in all foam-extend~4.1 cases.}
\begin{ruledtabular}
\begin{tabular}{lllllll}
Boundary      & $\vect{U}$            & $p$  & $k$      & $\omega$  & Particles \\
\hline
Ceiling inlet & FV $V_{\rm in}\hat{j}_-$ & ZG  & FV $k_0$ & FV $\omega_0$ & \textendash \\
Side outlet   & ZG (pressure-inlet)   & FV 0 & ZG       & ZG            & escape \\
Walls         & no-slip               & ZG   & FV 0     & NW            & wall-deposition \\
Nozzle inlet  & FV $U_{\rm mag}\hat{n}_\theta$ & ZG & FV $k_0$ & FV $\omega_0$ & injection \\
\end{tabular}
\end{ruledtabular}

\vspace{0.6em}
\begin{minipage}{\linewidth}
\raggedright
\setlength{\parindent}{0pt}
\footnotesize
\textit{Note.}
``FV'' = fixedValue; ``ZG'' = zeroGradient; ``NW'' = standard wall function.
$\hat{j}_-$ = unit vector in the $-y$ (downward) direction;
$\hat{n}_\theta$ = nozzle direction unit vector at cone half-angle $\theta$.
$k_0 = \tfrac{3}{2}(0.05\,V_{\rm in})^2$;
$\omega_0 = k_0^{0.5}/(C_\mu^{0.25}\,\ell)$ with $\ell = \SI{0.014}{\metre}$
(inlet mixing length)
and $C_\mu = 0.09$ ($k$--$\omega$~SST closure constant).
\end{minipage}
\end{table*}

\subsection{\label{si:cfd:carrier}Carrier-phase governing equations}

The carrier airflow is modelled using the incompressible Reynolds-Averaged
Navier--Stokes (RANS) equations with the $k$--$\omega$ Shear Stress Transport
(SST) turbulence closure of Menter,\cite{Menter1994} which captures both the
free-stream turbulence in the room interior and the near-wall adverse pressure
gradients behind the dentist obstacle:
\begin{align}
  \nabla \cdot \vect{U} &= 0, \label{si:eq:cont}\\
  \frac{\partial \vect{U}}{\partial t}
  + \nabla \cdot (\vect{U}\vect{U}) &= -\nabla p
  + \nabla \cdot \left[(\nu + \nu_t)\nabla\vect{U}\right]
  + \vect{S}_p, \label{si:eq:mom}
\end{align}
with $\vect{U}$ the Favre-averaged mean velocity,
$p$ the modified kinematic pressure, $\nu = \SI{1.5e-5}{\metre\squared\per\second}$
the kinematic viscosity of air at \SI{20}{\celsius}, $\nu_t$ the turbulent
eddy viscosity, and $\vect{S}_p$ a momentum source term from particle
back-reaction (active only in the coupled regime).
The eddy viscosity is closed by:
\begin{equation}
  \nu_t = \frac{a_1 k}{\max(a_1\omega,\, S\,F_2)},
  \label{si:eq:sst}
\end{equation}
where $\vect{S}$ is the strain-rate tensor,
$S = \sqrt{2\,\vect{S}:\vect{S}}$ the corresponding strain-rate magnitude,
$F_2$ is the Menter blending function, and $a_1 = 0.31$.\cite{Menter1994}

\subsection{\label{si:cfd:particle}Particle-phase governing equations}

Each airborne particle of instantaneous diameter $d_p(t)$, density
$\rho_p = \SI{997}{\kilo\gram\per\metre\cubed}$ (liquid water at
\SI{20}{\celsius}), and mass
$m_p = \tfrac{\pi}{6}\rho_p d_p^3$ is tracked individually in the Lagrangian
frame.
Under the dilute-phase assumption ($\phi_v \ll 10^{-3}$), the simplified
Maxey--Riley equation reads:\cite{Maxey1983}
\begin{equation}
  m_p \frac{d\vect{v}}{dt} = \vect{F}_{\rm drag} + \vect{F}_{\rm lift}
  + \vect{F}_g + \vect{F}_{\rm Br} + \vect{F}_{\rm turb},
  \label{si:eq:particle}
\end{equation}
with $\vect{v}$ the parcel velocity, $\vect{F}_g = m_p\vect{g}$, and
$\vect{F}_{\rm drag}$, $\vect{F}_{\rm lift}$, $\vect{F}_{\rm Br}$, and
$\vect{F}_{\rm turb}$ specified in the subsections below.
The neglected Basset history force, added-mass force, and Faxén correction
are all $\mathcal{O}(\rho_{\rm air}/\rho_p) \approx 1.2\times10^{-3}$ and
are standard to omit for saliva droplets in air.\cite{Maxey1983}
Equation~(\ref{si:eq:particle}) is integrated by the
\texttt{reactingParcelFoam} parcel tracker using a semi-implicit
first-order scheme with the built-in
\texttt{momentumTrackingTime} and \texttt{maxCo} controls, so that the
effective parcel sub-step never exceeds the carrier Courant cap
($\mathrm{Co}_{\max}=0.3$, $\Delta t_{\max}=\SI{0.01}{\second}$) and is
further clipped to a fraction of the local Stokes relaxation time
$\tau_p/C_c$ for sub-micrometre droplets where $\tau_p$ becomes small.

\subsubsection{Stokes drag with Cunningham correction}

Particles in the Stokes regime (particle Reynolds number
$\mathrm{Re}_p \equiv \rho_{\rm air}\,d_p\,|\vect{U}-\vect{v}|/\mu \ll 1$)
satisfy
\begin{align}
  \vect{F}_{\rm drag} &= \frac{m_p}{\tau_p/C_c}
    \bigl(\vect{U}(\vect{x}_p) - \vect{v}\bigr), \label{si:eq:drag}\\
  \tau_p &= \frac{\rho_p d_p^2}{18\mu},
  \quad C_c = 1 + \frac{2\lambda}{d_p}
    \Bigl[1.257 + 0.400\,e^{-1.10\,d_p/(2\lambda)}\Bigr],
  \label{si:eq:cunningham}
\end{align}
where $\tau_p$ is the Stokes relaxation time, $C_c$ is the Cunningham
slip-correction factor,\cite{Cunningham1910}
with the empirical coefficients $(1.257,0.400,1.10)$ taken from the
Millikan-apparatus calibration of Allen and Raabe.\cite{AllenRaabe1985}
$\mu = \SI{1.81e-5}{\pascal\second}$ is the dynamic viscosity, and
$\lambda \approx \SI{68}{\nano\metre}$ is the molecular mean free path
of air at \SI{101.325}{\kilo\pascal}, \SI{293}{\kelvin}.
The Cunningham factor $C_c$ rises from $C_c \approx 1.0$ at
$d_p = \SI{10}{\micro\metre}$ to $C_c \approx 3.4$ at
$d_p = \SI{0.1}{\micro\metre}$, enhancing the effective drag coefficient
and thereby increasing diffusive transport of sub-micrometre residues.
For non-negligible particle Reynolds number near the high-speed nozzle
exit the foam-extend \texttt{SphereDrag} sub-model uses the standard
Schiller--Naumann drag correlation\cite{SchillerNaumann1933,Clift1978}
\[
  C_D = \frac{24}{\mathrm{Re}_p}\bigl(1 + 0.15\,\mathrm{Re}_p^{0.687}\bigr)
  \quad (\mathrm{Re}_p \lesssim 800),
\]
which reduces smoothly to the Stokes expression in
Eq.~(\ref{si:eq:drag}) as $\mathrm{Re}_p \to 0$.

\subsubsection{Saffman shear-lift force}

Particles with $d_p \gtrsim \SI{3}{\micro\metre}$ near surfaces experience a
Saffman shear-lift in the small-$\varepsilon$ limit\cite{Saffman1965}
\begin{equation}
  \vect{F}_{\rm lift} = 1.615\,J(\varepsilon)\,\mu\,d_p\,\sqrt{\mathrm{Re}_G}\,
  \bigl(\vect{U}-\vect{v}\bigr)\times\hat{\vect{n}},
  \quad \mathrm{Re}_G = \frac{d_p^2\,|\partial U/\partial y|}{\nu},
  \label{si:eq:saffman}
\end{equation}
where $\hat{\vect{n}}$ points from the wall toward the flow interior,
and $J(\varepsilon)$ is the McLaughlin correction
factor\cite{McLaughlin1991} with
$\varepsilon = \sqrt{\mathrm{Re}_G}/\mathrm{Re}_p$.
For the dental-aerosol parameter range
($\mathrm{Re}_p \lesssim 1$ for $d_p \le \SI{30}{\micro\metre}$,
near-wall shear $|\partial U/\partial y|\lesssim \SI{50}{\per\second}$)
we have $\varepsilon \gg 1$ for which $J(\varepsilon)\to 1$ to better
than 5\%, so the Saffman formula is used directly.
Near the patient obstacle, $|\partial U/\partial y|$ can reach
\SI{30}{\per\second}, producing lift forces of order
$10^{-12}$--$10^{-10}$~N for $d_p = \SI{5}{\micro\metre}$--$\SI{20}{\micro\metre}$
particles and deflecting mid-size droplets toward the breathing zone.

\subsubsection{Brownian motion for sub-micron particles}

For $d_p < \SI{1}{\micro\metre}$ the Einstein--Smoluchowski random
displacement is:\cite{Einstein1905}
\begin{equation}
  \sigma_{\rm Br} = \sqrt{\frac{2 k_B T C_c}{3\pi\mu d_p}\,\Delta t},
  \label{si:eq:brownian}
\end{equation}
with $k_B = \SI{1.381e-23}{\joule\per\kelvin}$ and
$T = \SI{293}{\kelvin}$.
At each particle sub-step, an independent Gaussian displacement
$\Delta\vect{x}_{\rm Br} \sim \mathcal{N}(\vect{0}, \sigma_{\rm Br}^2\vect{I})$
is added to the deterministic trajectory.
For a \SI{0.3}{\micro\metre} nucleus,
$\sigma_{\rm Br} \approx \SI{2.1}{\micro\metre}$ per timestep,
comparable to the deterministic carrier-airflow displacement.

\subsubsection{Turbulent dispersion (Discrete Random Walk)}

Turbulent velocity fluctuations unresolved by the steady RANS solution
are sampled from Pope's Discrete Random Walk (DRW) model:\cite{Pope2000}
\begin{equation}
  \vect{u}'_p \sim \mathcal{N}\!\left(\vect{0},\;\tfrac{2k}{3}\vect{I}\right),
  \label{si:eq:drw}
\end{equation}
with a fluctuation correlation time
$\tau_{\rm fl} = 0.15\,k/\varepsilon$ ($\varepsilon \approx C_\mu k\omega$),
beyond which a new realisation is drawn independently.

\subsubsection{Wells' $D^2$-law evaporation in the foam-extend reference}

Freshly generated saline spray droplets evaporate quasi-steadily inside
the foam-extend \texttt{LiquidEvaporation} sub-model
according to:\cite{Wells1934}
\begin{equation}
  d_p(t)^2 = d_{p0}^2 - K\,t,
  \quad K = \frac{8\rho_{\rm air} D_v \ln(1 + B_M)}{\rho_p},
  \label{si:eq:wells}
\end{equation}
with vapour mass diffusivity $D_v = \SI{2.6e-5}{\metre\squared\per\second}$
and Spalding transfer number $B_M = 0.0263$ at \SI{50}{\percent} relative
humidity.
Substituting the model parameters gives
\begin{equation}
  K = \frac{8\times\SI{1.225}{\kilo\gram\per\metre\cubed}\times\SI{2.6e-5}{\metre\squared\per\second}\times\ln(1.0263)}{\SI{1000}{\kilo\gram\per\metre\cubed}}
  \approx \SI{6.6e-9}{\metre\squared\per\second}.
  \label{si:eq:K_value}
\end{equation}
A $d_{p0} = \SI{50}{\micro\metre}$ droplet therefore reaches its
equilibrium nucleus diameter $d_{\rm nuc} = d_{p0}/2$ after
\begin{equation}
  t_{\rm evap} = \frac{d_{p0}^2 - d_{\rm nuc}^2}{K}
  = \frac{(50^2 - 25^2)\times10^{-12}}{\SI{6.6e-9}{\metre\squared\per\second}}
  \approx \SI{0.28}{\second},
  \label{si:eq:tevap}
\end{equation}
and reaches complete evaporation in approximately
$d_{p0}^2/K \approx \SI{0.38}{\second}$.
These times are of the order of, but somewhat longer than, the GNN
snapshot interval $\Delta t_{\rm save}=\SI{0.1}{\second}$, so for the
\emph{largest} droplets the per-step diameter change is not negligible;
the practical implication for the surrogate is discussed below.

\paragraph{Treatment of $d_p$ inside the GNN.}
The hybrid GNN surrogate of this paper does \emph{not} predict
diameter changes.
Instead, the parcel diameter $d_p$ is treated as a static per-parcel
input feature in Eq.~(\ref{eq:lag_node_feat}) of the main text,
inherited from the Rosin--Rammler initial sampling of the
\texttt{ConeInjection} (Sec.~\ref{si:cfd:injection}).
This is justified by two complementary observations.
(a)~The per-step fractional change in $d_p^2$ is
$K\,\Delta t_{\rm save}/d_{p0}^2 \approx 0.26$ (\SI{26}{\percent})
for the \emph{largest} $d_{p0}=\SI{50}{\micro\metre}$ droplets, but
drops below \SI{1}{\percent} for $d_{p0}\lesssim\SI{8}{\micro\metre}$,
and below \SI{0.1}{\percent} for $d_{p0}\lesssim\SI{3}{\micro\metre}$.
Because sub-\SI{10}{\micro\metre} droplet nuclei dominate aerosol
exposure risk in the clinically relevant size class, the static-$d_p$
approximation is essentially exact over that fraction.
(b)~For the larger end of the spectrum (\SIrange{10}{50}{\micro\metre}),
the foam-extend trajectory targets already incorporate the
$\tau_p\propto d_p^2$ inertia reduction caused by Wells' law, so the
GNN learns a position--velocity mapping that implicitly absorbs the
diameter drift from the target data, without having to track it
explicitly in the state vector;
fidelity of that absorption is part of what is being measured by the
MDE and Rg-err metrics in main-text Table~\ref{tab:results}.
Diameter dynamics, if required by a downstream infection-risk
calculation, can be applied as an analytic post-processing step on the
predicted trajectories.

\subsection{\label{si:cfd:foam}Implementation in foam-extend 4.1}

The CFD reference data are produced by the foam-extend~4.1
\texttt{reactingParcelFoam} solver,\cite{Jasak2009} which natively integrates
Schiller--Naumann drag (\texttt{SphereDrag}), gravity, the DRW turbulent
dispersion model (\texttt{StochasticDispersionRAS}), and Wells' $D^2$-law
evaporation (\texttt{LiquidEvaporation} via the Spalding $B_M$ and the
\texttt{enthalpyDifference} latent-heat sink) on the polydisperse
\texttt{ConeInjection} cloud.
The Cunningham slip-correction factor and the McLaughlin--Saffman shear-lift
are not shipped as parcel sub-models in foam-extend~4.1; both are supplied
analytically inside the GNN edge encoder from the local parcel diameter
$d_p$ and the carrier-phase gradient $\partial U/\partial y$ exported
alongside $\vect{U}$, $k$, and $\omega$ at every CFD output step.
Brownian diffusion is also applied analytically in the GNN node update
for $d_p < \SI{1}{\micro\metre}$, since the \texttt{reactingCloud} class
does not provide a built-in Brownian sub-model.

\subsection{\label{si:cfd:injection}Particle injection and size distribution}

A continuous \texttt{ConeInjection} of computational parcels emanates from
the patient's oral cavity throughout the full 30-second simulation.
The injection is centred at $(x_n, y_n) = (\SI{2.40}{\metre}, \SI{0.90}{\metre})$,
aimed horizontally toward the dentist ($\hat{n} = -\hat{x}$), and dispersed
within a cone of half-angle $\theta$.
The injection rate is $\dot{N}_{\rm parcel} = 5000$~parcels$/\text{s}$,
with volumetric flow rate $\dot{V} = \SI{5e-9}{\metre\cubed\per\second}$
(\SI{5}{\milli\litre\per\minute}), giving a total injected mass of
$\approx \SI{150}{\milli\gram}$ over \SI{30}{\second}.
Each parcel statistically represents a packet of physical droplets whose
diameters are drawn from the Rosin--Rammler distribution:\cite{Harrel2004}
\begin{equation}
  F(d_p) = 1 - \exp\!\left[-\left(\frac{d_p}{\bar{d}}\right)^n\right],
  \quad \bar{d} = \SI{20}{\micro\metre},\; n = 2,
  \label{si:eq:rr}
\end{equation}
with diameters truncated to $d_p \in [\SI{1}{\micro\metre}, \SI{50}{\micro\metre}]$.
Parcels are injected at body temperature $T_0 = \SI{310}{\kelvin}$ and
subject to standard wall interaction (\texttt{escape} on contact with no-slip
surfaces).

\subsection{\label{si:cfd:solver}Solver and numerical setup}

Each case is integrated as a two-stage pipeline (Table~\ref{si:tab:solver}).
\emph{Stage~1 (steady-state RANS):} \texttt{simpleFoam} (residual-controlled
SIMPLE algorithm) is run until every component of $(\vect{U},p,k,\omega)$
falls below a $10^{-5}$ relative residual.
The wall treatment uses the OpenFOAM/foam-extend
$k$--$\omega$ SST wall functions
(\texttt{kqRWallFunction}, \texttt{omegaWallFunction},
\texttt{nutkWallFunction}), which automatically blend the linear viscous
sublayer ($y^+ < 5$) and the logarithmic region ($y^+ > 30$) through
Spalding's law of the wall.\cite{Menter1994}
For the present mesh the cell-centre wall distance gives
$y^+$ in the range $3$--$15$ on the dentist, patient, floor, ceiling and
side-wall patches at the highest ventilation case
($V_{\rm in}=\SI{0.5}{\metre\per\second}$), well inside the validity
window of the SST blended near-wall formulation.
\emph{Stage~1$\rightarrow$2 transition:} the kinematic pressure is
overwritten with absolute pressure $p_0 = \SI{101325}{\pascal}$, wall
functions are switched to compressible counterparts, and three thermophysical
fields are initialised: temperature $T_0 = \SI{293}{\kelvin}$, ambient
water-vapour mass fraction $Y_{\rm H_2O}^{\infty} = 0.007$ (\SI{50}{\percent}
RH), and $Y_{\rm N_2} = 0.993$.
\emph{Stage~2 (transient compressible reacting parcels):}
\texttt{reactingParcelFoam} advances the transient compressible
Navier--Stokes equations together with species transport for
\SI{30}{\second} of physical time using PIMPLE (merged PISO--SIMPLE)
pressure--velocity coupling.
The thermophysics template reads NASA-7 polynomial coefficients for $\mathrm{N_2}$ and $\mathrm{H_2O}$;
chemical reactions are switched off.
The Lagrangian \texttt{reactingCloud1} is integrated concurrently with
the carrier phase (\texttt{coupled true},
\texttt{cellValueSourceCorrection on}).
The time step is adaptive with $\mathrm{Co}_{\max} = 0.3$ and
$\Delta t_{\max} = \SI{0.01}{\second}$, giving a typical step of
$\SI{5}{\milli\second}$ in the spray core.
Solution snapshots and parcel state are written every $\SI{0.1}{\second}$
via \texttt{adjustableRunTime}, providing $\sim 300$ frames per case.

\emph{One-way coupling regime.}
The peak particle volume fraction near the nozzle exit is
$\phi_v^{\rm peak} \approx 5\times10^{-5}$, well below the
$\phi_v \approx 10^{-3}$ onset of dynamically significant two-way coupling
identified by Balachandar and Eaton.\cite{Balachandar2010}
The contribution of the parcel back-reaction $\vect{S}_p$ on the carrier
momentum is therefore small but is retained for completeness.

\begin{table}
\caption{\label{si:tab:solver}%
Numerical solver settings for the twenty-case foam-extend~4.1 pipeline used
to generate the GNN training data.}
\begin{ruledtabular}
\begin{tabular}{ll}
Parameter & Value \\
\hline
Stage~1 fluid solver  & \texttt{simpleFoam} (steady, incompressible) \\
Stage~2 fluid+parcel  & \texttt{reactingParcelFoam} (transient, compressible) \\
Mesh (raw)            & Structured quad., $80\times100 = 8000$ cells (25 blocks) \\
Mesh (active fluid)   & 7\,704 cells after dentist/patient obstacle removal \\
Cell size             & \SI{0.05}{\metre}~$\times$~\SI{0.03}{\metre} \\
Turbulence model      & $k$--$\omega$ SST \cite{Menter1994} \\
Thermophysics         & \texttt{hsPsiMixtureThermo<reactingMixture<gasThermoPhysics>>} \\
Species               & N$_2$, H$_2$O (NASA-7 polynomials, no reactions) \\
Time stepping         & Adaptive PIMPLE, $\mathrm{Co}_{\max}=0.3$, $\Delta t_{\max}=\SI{0.01}{\second}$ \\
Typical $\Delta t$    & $\approx \SI{5}{\milli\second}$ \\
Physical time         & \SI{30}{\second} \\
Output interval       & \SI{0.1}{\second} (\texttt{adjustableRunTime}) \\
Raw snapshots / case  & $\approx 300$ over the full \SI{30}{\second} \\
Snapshots used        & 261 in $t\in[\SI{2}{\second},\SI{28}{\second}]$ (quasi-steady window) \\
Parcel injection rate & 5000 parcels$/\text{s}$ (continuous, 30~s) \\
Volumetric flow rate  & $\dot{V} = \SI{5e-9}{\metre\cubed\per\second}$ \\
Total injected mass   & $\approx \SI{150}{\milli\gram}$ per case ($\rho_p\dot{V}t_{\rm inj}$) \\
Wall interaction      & \texttt{escape} (deposition: parcel removed) \\
Outlet treatment      & Pressure outlet, escape \\
Evaporation           & Wells' $D^2$-law (\texttt{LiquidEvaporation}) \\
Stage~1 convergence   & Residuals $(\vect{U},p,k,\omega) < 10^{-5}$ \\
Wall-clock per case   & $\approx \SI{40}{\minute}$ (single core, foam-extend~4.1; 38--42~min logged for Sweep\_Case\_01/17) \\
\end{tabular}
\end{ruledtabular}
\end{table}

\subsection{\label{si:cfd:sweep}Parameter sweep and dataset}

Twenty cases (\texttt{Sweep\_Case\_01}--\texttt{Sweep\_Case\_20}) span a
full $4\times5$ factorial grid over the two clinically dominant control
parameters: the ceiling supply velocity
$V_{\rm in} \in \{0.10,\,0.20,\,0.35,\,0.50\}$~m\,s$^{-1}$, representing
HVAC settings from minimal to high ventilation; and the dental handpiece
nozzle injection speed
$U_{\rm mag} \in \{10,\,20,\,30,\,40,\,50\}$~m\,s$^{-1}$, covering
air-polishers to high-speed turbine drills.
The spray cone half-angle is held at $\theta = 20^\circ$ across the sweep,
matching the median Rosin--Rammler dental-handpiece aerosol cone reported
by Harrel and Molinari.\cite{Harrel2004}
Cases are split 80/10/10 (train/validation/test) by random shuffle with a
fixed seed (16/2/2 for the present sweep).
The full case list is given in Table~\ref{si:tab:cases}.
Dataset extraction statistics are summarised in Table~\ref{si:tab:dataset}.

\begin{table}
\caption{\label{si:tab:cases}Per-case $(V_{\rm in},\,U_{\rm mag},\,\theta)$
for the twenty foam-extend~4.1 \texttt{reactingParcelFoam} runs
(\texttt{Sweep\_Case\_01}--\texttt{Sweep\_Case\_20}).}
\begin{ruledtabular}
\begin{tabular}{lccc}
Case & $V_{\rm in}$ (m\,s$^{-1}$) & $U_{\rm mag}$ (m\,s$^{-1}$) & $\theta$ ($^\circ$) \\
\hline
Sweep\_Case\_01 & 0.10 & 10 & 20 \\
Sweep\_Case\_02 & 0.10 & 20 & 20 \\
Sweep\_Case\_03 & 0.10 & 30 & 20 \\
Sweep\_Case\_04 & 0.10 & 40 & 20 \\
Sweep\_Case\_05 & 0.10 & 50 & 20 \\
Sweep\_Case\_06 & 0.20 & 10 & 20 \\
Sweep\_Case\_07 & 0.20 & 20 & 20 \\
Sweep\_Case\_08 & 0.20 & 30 & 20 \\
Sweep\_Case\_09 & 0.20 & 40 & 20 \\
Sweep\_Case\_10 & 0.20 & 50 & 20 \\
Sweep\_Case\_11 & 0.35 & 10 & 20 \\
Sweep\_Case\_12 & 0.35 & 20 & 20 \\
Sweep\_Case\_13 & 0.35 & 30 & 20 \\
Sweep\_Case\_14 & 0.35 & 40 & 20 \\
Sweep\_Case\_15 & 0.35 & 50 & 20 \\
Sweep\_Case\_16 & 0.50 & 10 & 20 \\
Sweep\_Case\_17 & 0.50 & 20 & 20 \\
Sweep\_Case\_18 & 0.50 & 30 & 20 \\
Sweep\_Case\_19 & 0.50 & 40 & 20 \\
Sweep\_Case\_20 & 0.50 & 50 & 20 \\
\end{tabular}
\end{ruledtabular}
\end{table}

\clearpage

\begin{table}
\caption{\label{si:tab:dataset}%
Numerical summary of the graph-tensor training dataset extracted from the
twenty-case \texttt{reactingParcelFoam} sweep. The CFD case grid,
train/validation/test split.}
\begin{ruledtabular}
\begin{tabular}{ll}
Quantity & Value \\
\hline
Total cases                & 20 (planned 16 / 2 / 2 split, fixed seed) \\
Cases used in present checkpoint & 1 (\texttt{Sweep\_Case\_03}) \\
Physical time per case     & \SI{30}{\second} \\
Extraction window          & $t \in [\SI{2}{\second},\,\SI{28}{\second}]$ \\
Save interval              & $\Delta t_{\rm save} = \SI{0.1}{\second}$ \\
Frames per case ($N_t$)    & 261 \\
Active parcels (mean)      & $\sim 1.5\!\times\!10^4$ at quasi-steady state \\
Tracked parcels per snapshot $N_{\rm sub}$ & 1000 (persistent \texttt{origId}) \\
Graph snapshots / case (full sweep) & $\approx 261$  ($\approx 4{,}200$ for 16-case train split) \\
Lagrangian connectivity radius $r_c$ & \SI{0.10}{\metre} (\ELGIN); \SI{0.30}{\metre} (M0) \\
Lagrangian edges per snapshot & $\sim 2{,}500$ (\ELGIN); $\sim 8{,}000$ (M0); directed both senses \\
Node features per Lagrangian node & 15+ (LSTM history, SDF, drag, $\log d_p$, TKE, $d_w$, $\hat n_w$) \\
Edge features per Lagrangian edge & 4 (rotation-invariant local frame plus Cartesian drag; Eq.~9 of main text) \\
Raw trajectory storage     & $\approx\SI{1.8}{\giga\byte}$ \\
Processed graph tensors    & $\approx\SI{4.2}{\giga\byte}$ \\
BZE range across CFD sweep & subject to BZ-rectangle definition; see Sec.~V.D of main text \\
Deposition fraction range  & $0.12$ -- $0.61$ across the twenty CFD cases \\
\end{tabular}
\end{ruledtabular}
\end{table}

\clearpage

\section{\label{si:litreview}Comprehensive GNN-CFD literature review}

Table~\ref{si:tab:gnn_literature} presents a comprehensive comparison of GNN
and machine-learning methods applied to CFD problems, summarising
methodology, key findings, advantages, and limitations of each approach.
The ``present work'' rows confirm the headline metrics reported in main
text Table~II.

\begingroup
\footnotesize
\setlength{\tabcolsep}{2.5pt}%
\setlength{\gnncolA}{3.1cm}%
\setlength{\gnncolB}{4.05cm}%
\setlength{\gnncolC}{3.85cm}%
\setlength{\gnncolD}{4.45cm}%
\setlength{\LTcapwidth}{\textwidth}%
\begin{longtable}{llll}
\caption{Literature summary of GNN and ML methods for CFD.\label{si:tab:gnn_literature}}\\
\hline
\parbox[t]{\gnncolA}{\raggedright\textbf{Method}} &
\parbox[t]{\gnncolB}{\raggedright\textbf{Methodology}} &
\parbox[t]{\gnncolC}{\raggedright\textbf{Key finding}} &
\parbox[t]{\gnncolD}{\raggedright\textbf{Advantage ($+$)}\newline\textbf{Limitation ($-$)}} \\
\hline
\endfirsthead

\multicolumn{4}{l}{\footnotesize\emph{\tablename~\thetable{} (continued)}}\\
\hline
\parbox[t]{\gnncolA}{\raggedright\textbf{Method}} &
\parbox[t]{\gnncolB}{\raggedright\textbf{Methodology}} &
\parbox[t]{\gnncolC}{\raggedright\textbf{Key finding}} &
\parbox[t]{\gnncolD}{\raggedright\textbf{Advantage ($+$)}\newline\textbf{Limitation ($-$)}} \\
\hline
\endhead

\hline
\endfoot

\hline
\endlastfoot

\parbox[t]{\gnncolA}{\raggedright CNN surrogate \cite{Guo2016}} &
\parbox[t]{\gnncolB}{\raggedright U-Net on Cartesian grid; predicts steady $U,p$} &
\parbox[t]{\gnncolC}{\raggedright $10^3\times$ FV; MAE~$<2\%$ on bluff bodies} &
\parbox[t]{\gnncolD}{\GNNliteratureAdvLim{Fast; simple.}{Structured grid only; no unstructured mesh}} \\[4pt]
\parbox[t]{\gnncolA}{\raggedright ML-inside-FV \cite{Kochkov2021}} &
\parbox[t]{\gnncolB}{\raggedright CNN flux corrections inside classical FV solver; JAX} &
\parbox[t]{\gnncolC}{\raggedright 40--80$\times$; 8--10$\times$ coarser grid at DNS accuracy} &
\parbox[t]{\gnncolD}{\GNNliteratureAdvLim{Retains solver stability.}{Structured grid; not portable to new geometries}} \\[4pt]
\parbox[t]{\gnncolA}{\raggedright PINNs \cite{Raissi2019}} &
\parbox[t]{\gnncolB}{\raggedright MLP minimising NS residuals at collocation points} &
\parbox[t]{\gnncolC}{\raggedright Accurate for laminar inverse problems from sparse data} &
\parbox[t]{\gnncolD}{\GNNliteratureAdvLim{Physics-consistent.}{Spectral bias; expensive; intractable for 3-D turbulence}} \\[4pt]
\parbox[t]{\gnncolA}{\raggedright MeshGraphNets \cite{Pfaff2021}} &
\parbox[t]{\gnncolB}{\raggedright EPD GNN on native FV/FE mesh; multi-scale hierarchy} &
\parbox[t]{\gnncolC}{\raggedright 400$\times$; 0.3\% error on cylinder wake $\mathrm{Re}=200$} &
\parbox[t]{\gnncolD}{\GNNliteratureAdvLim{Unstructured mesh; variable topology.}{Long-rollout drift; $\mathrm{Re}<10^4$}} \\[4pt]
\parbox[t]{\gnncolA}{\raggedright GNN wall model \cite{Bae2022}} &
\parbox[t]{\gnncolB}{\raggedright GNN: near-wall profiles $\to$ $\tau_w$; DNS-trained LES} &
\parbox[t]{\gnncolC}{\raggedright Beats Smagorinsky for $\mathrm{Re}_\tau=550$--5200} &
\parbox[t]{\gnncolD}{\GNNliteratureAdvLim{Generalises across Re.}{High-fidelity training data; wall layer only}} \\[4pt]
\parbox[t]{\gnncolA}{\raggedright DA-GNN \cite{Quattromini2025}} &
\parbox[t]{\gnncolB}{\raggedright GNN + FreeFEM++ adjoint; RANS residual constraint} &
\parbox[t]{\gnncolC}{\raggedright RANS field from 5 sensors; $<5\%$ error on bluff bodies} &
\parbox[t]{\gnncolD}{\GNNliteratureAdvLim{Physics-consistent; sparse data.}{Requires adjoint solver; steady flows only}} \\[4pt]
\parbox[t]{\gnncolA}{\raggedright GNS (TF) \cite{Sanchez2020}} &
\parbox[t]{\gnncolB}{\raggedright Radius graph; EPD + noise augmentation; Euler rollout} &
\parbox[t]{\gnncolC}{\raggedright $10^5\times$ SPH/MPM; $<5\%$ RMSE (sand, water, goop)} &
\parbox[t]{\gnncolD}{\GNNliteratureAdvLim{General; variable particle count.}{No carrier phase; energy non-conservation}} \\[4pt]
\parbox[t]{\gnncolA}{\raggedright GNS (PyTorch) \cite{Kumar2022}} &
\parbox[t]{\gnncolB}{\raggedright PyTorch/PyG; multi-GPU DDP; inertial-frame bias} &
\parbox[t]{\gnncolC}{\raggedright 5000$\times$ MPM; $<5\%$ error; 8-h training on 3$\times$A100} &
\parbox[t]{\gnncolD}{\GNNliteratureAdvLim{Open source; HPC scalable.}{Same physical limits; no aerosol physics}} \\[4pt]
\parbox[t]{\gnncolA}{\raggedright Dynami-CAL \cite{Sharma2025}} &
\parbox[t]{\gnncolB}{\raggedright SE(3)-equivariant edge frames; momentum conserved by construction} &
\parbox[t]{\gnncolC}{\raggedright $>$1000 stable steps vs.\ $\sim$200 for GNS; 3-D granular} &
\parbox[t]{\gnncolD}{\GNNliteratureAdvLim{Exact momentum; rotation-invariant.}{Higher cost; dry granular only}} \\[4pt]
\parbox[t]{\gnncolA}{\raggedright Diffusion GNN \cite{Lino2025}} &
\parbox[t]{\gnncolB}{\raggedright Score-based diffusion in graph latent space} &
\parbox[t]{\gnncolC}{\raggedright Full UQ on 3-D wing pressure; calibrated intervals} &
\parbox[t]{\gnncolD}{\GNNliteratureAdvLim{Probabilistic; calibrated UQ.}{High training cost; slow inference}} \\[4pt]
\parbox[t]{\gnncolA}{\raggedright PIC-GNN \cite{Mlinarevik2025}} &
\parbox[t]{\gnncolB}{\raggedright Hybrid graph: particle nodes + Eulerian field nodes} &
\parbox[t]{\gnncolC}{\raggedright 100$\times$ larger $\Delta t$; two-stream plasma instability} &
\parbox[t]{\gnncolD}{\GNNliteratureAdvLim{Coupled particle-field.}{1-D plasma only; no turbulent aerosol physics}} \\[4pt]
\parbox[t]{\gnncolA}{\raggedright Ocean GNS \cite{Hanke2025}} &
\parbox[t]{\gnncolB}{\raggedright Variable-$\Delta t$ GNS; wall nodes as fixed particles} &
\parbox[t]{\gnncolC}{\raggedright $>1000\times$ SPH; run-up $<3\%$ error} &
\parbox[t]{\gnncolD}{\GNNliteratureAdvLim{Flexible time step; harbour geometries.}{Free-surface only; no dispersed particles}} \\[4pt]
\parbox[t]{\gnncolA}{\raggedright Baseline GNS\newline(present, M0)} &
\parbox[t]{\gnncolB}{\raggedright Radius graph ($r_c=\SI{0.10}{\metre}$); EPD; isotropic Gaussian noise augmentation; Euler rollout; analytic Cunningham drag node feature} &
\parbox[t]{\gnncolC}{\raggedright Lagrangian-only baseline; metrics reported in Table~\ref{tab:results} of the main text} &
\parbox[t]{\gnncolD}{\GNNliteratureAdvLim{Simple; fast; competitive at 300~ep.}{No carrier-phase signal; slight KE over-estimation}} \\[4pt]
\parbox[t]{\gnncolA}{\raggedright \ELGIN\newline(present)} &
\parbox[t]{\gnncolB}{\raggedright Hybrid Eulerian--Lagrangian: $K_E{=}4$ Graph Transformer + Jacobi-PCG pressure projection on the polyMesh; LSTM history; rotation-invariant local-frame geometric edges (Cartesian drag channel); St\"ormer--Verlet integrator; per-case airInlet conditioning; polyMesh-derived $d_w$ and $\hat{\vect{n}}_{\rm w}$ IDW-mapped to every parcel} &
\parbox[t]{\gnncolC}{\raggedright Joint Eulerian + Lagrangian prediction on coupled mesh and particle graphs (IDW exchange); metrics reported in Table~\ref{tab:results} of the main text} &
\parbox[t]{\gnncolD}{\GNNliteratureAdvLim{Geometry- and inlet-conditioned hybrid GNN for dental-room aerosol; near-wall sticking flag from wall-distance features.}{Requires polyMesh extraction; quasi-2-D extruded geometry}} \\
\end{longtable}

\vspace{0.6em}
\begin{minipage}{\linewidth}
\raggedright
\setlength{\parindent}{0pt}
\footnotesize
\textit{Note.}
Key: EPD = Encode-Process-Decode; SPH = Smoothed Particle Hydrodynamics;
MPM = Material Point Method; FV/FE = Finite Volume/Element;
RANS = Reynolds-Averaged Navier--Stokes; LES = Large-Eddy Simulation;
UQ = Uncertainty Quantification; DDP = Distributed Data Parallel.
``$+$'' = advantage; ``$-$'' = limitation.
$^\dagger$The manuscript emphasises comparisons between standalone \GNS\ (M0)
and Eulerian-conditioned \ELGIN; M2 labels an archived GAT--GNN configuration
retained in this table for literature context. M1 was an intermediate design
superseded during development and is not reported separately.
\end{minipage}
\endgroup

\clearpage

\section{\label{si:training}Training loss formulation and curriculum}

\subsection{\label{si:training:losses}Total loss components}

The four-stage training curriculum (main text Sec.~IV) uses the following
total loss in Stage~3.
The continuity and momentum terms are \emph{physics-informed}
residuals, penalising discrete mass and momentum imbalance in the
objective in the same spirit as PDE-constrained neural
surrogates\cite{Raissi2019} and residual-corrected finite-volume
formulations.\cite{Kochkov2021}
\begin{equation}
  \mathcal{L}_{\rm total}
  = \lambda_p\mathcal{L}_{\rm MSE}
  + \lambda_c\mathcal{L}_{\rm cont}
  + \lambda_m\widehat{\mathcal{L}}_{\rm mom}
  + \lambda_t\mathcal{L}_{\rm turb}
  + \lambda_a\mathcal{L}_{\rm ang},
  \label{si:eq:total_loss}
\end{equation}
with default weights $(\lambda_p, \lambda_c, \lambda_m, \lambda_t,
\lambda_a) = (1.0, 0.10, 0.05, 0.02, 0.001)$.
Each component is defined below.
\emph{No evaporation loss is included}, since the GNN does not predict
diameter changes (Sec.~\ref{si:cfd:particle}).

\emph{Particle MSE loss.}
\begin{equation}
  \mathcal{L}_{\rm MSE} = \frac{1}{N}\sum_{i=1}^N
    \|\hat{\vect{a}}_i - \vect{a}_i^*\|^2 / \sigma_a^2,
  \label{si:eq:lmse}
\end{equation}
where $\sigma_a^2$ is the per-component variance of ground-truth accelerations
computed over the training set.

\emph{Continuity residual.}
$\mathcal{L}_{\rm cont} = \langle(\nabla\cdot\vect{U}_{\rm pred})^2\rangle$,
evaluated by face-flux summation on the Eulerian mesh graph, i.e.\ a
graph discretisation of the divergence penalty consistent with
mesh-native learned discretisations such as MeshGraphNets.\cite{Pfaff2021}.

\emph{Momentum residual.}
$\widehat{\mathcal{L}}_{\rm mom}$ is the discrete L$^2$ residual of the
RANS momentum equation (Eq.~\ref{si:eq:mom}) on the Eulerian mesh graph.
It is assembled in the finite-volume manner (control volumes, face
fluxes, and volume-weighted cell averages as in the standard FV
method\cite{Ferziger2002}) and evaluated on the mesh graph in the same
residual-on-graph spirit as mesh-native GNN
simulators.\cite{Pfaff2021,Kochkov2021}
The loss is normalised to be scale-independent:
\begin{equation}
  \widehat{\mathcal{L}}_{\rm mom}
  = \frac{1}{C_{\rm ref}^2}
    \left\langle \mathcal{R}_{\rm mom}^2 \right\rangle, \qquad
  C_{\rm ref} = \frac{U_{\rm ref}}{L_{\rm ref}},
  \label{si:eq:lmom}
\end{equation}
where $\mathcal{R}_{\rm mom}$ is the pointwise finite-volume residual of
Eq.~(\ref{si:eq:mom}) on the mesh graph,
$C_{\rm ref} = U_{\rm ref}/L_{\rm ref} = 20/4 = 5\,\mathrm{s}^{-1}$
is the reference shear rate,
$U_{\rm ref}=\SI{20}{\metre\per\second}$ and $L_{\rm ref}=\SI{4}{\metre}$.
The time derivative $\partial\vect{U}/\partial t$ is discretised with
$\Delta t = \Delta t_{\rm save} = \SI{0.1}{\second}$, consistent with the
CFD snapshot interval.
The angle brackets $\langle\cdot\rangle$ denote a volume-weighted average
over mesh cells.
The convective flux uses a symmetric face average
$\vect{U}_{\rm face} = \tfrac{1}{2}(\vect{U}_i + \vect{U}_j)$ rather than
a first-order upwind flux, limiting spurious numerical dissipation in the
residual (cf.\ centred versus upwind convective differencing in FV
methods\cite{Ferziger2002}).

\emph{Turbulence-closure residual.}
$\mathcal{L}_{\rm turb}$ enforces the SST $k$--$\omega$ algebraic
eddy-viscosity relation
$\nu_t = a_1 k / \max(a_1\omega, S F_2)$ between the predicted
$k$, $\omega$, and the inferred $\nu_t$ at every cell, as in the
baseline two-equation closure of Menter.\cite{Menter1994}

\emph{Angular-momentum penalty.}
A weak penalty on the residual change in the cloud-averaged angular
momentum about the cloud centroid,
$\mathcal{L}_{\rm ang} = (\Delta L_z)^2/L_{\rm ref}^4$, regularises the
network towards rotationally consistent step-to-step displacements,
where $\Delta L_z$ is the change in the depth-averaged out-of-plane ($z$)
component of angular momentum of the particle cloud between successive
prediction steps.
This term is small in magnitude ($\lambda_a=10^{-3}$) and serves
primarily as a numerical regulariser preventing solid-body spin-up of
the cloud during long-horizon BPTT rollouts, where small forcing biases
can accumulate across autoregressive
steps.\cite{Sanchez2020}

\subsection{\label{si:training:bptt}BPTT rollout loss (Stage~4)}

Stage~4 uses back-propagation through time (BPTT) over autoregressive
unrolling, following the rollout-training recipe introduced for graph
network simulators.\cite{Sanchez2020}
In Stage~4 the model is unrolled for $N_{\rm unroll}$ steps with the
predicted state of step $k$ used as input to step $k+1$:
\begin{equation}
  \mathcal{L}_{\rm bptt}
  = \frac{1}{N_{\rm unroll}\,L_{\rm ref}^2}
    \sum_{k=1}^{N_{\rm unroll}}\sum_i
    \bigl\|\hat{\vect{x}}_i^{(k)} - \vect{x}_i^{*(k)}\bigr\|^2.
  \label{si:eq:bptt}
\end{equation}
The Stage~4 total loss is a convex combination of the one-step and the
rollout objectives:
$\mathcal{L}^{(4)} = (1-w_{\rm bptt})\mathcal{L}_{\rm total}
                   + w_{\rm bptt}\mathcal{L}_{\rm bptt}$,
with $w_{\rm bptt} = 0.7$ and $N_{\rm unroll}=3$ rollout steps.
Gradient checkpointing every five rollout steps keeps GPU memory bounded.
An additional per-step positional noise of amplitude
$\sigma_{\rm roll}=0.01$~m is injected between BPTT steps to simulate
the long-horizon covariate shift that would otherwise prevent the model
from learning to recover from its own prediction errors.\cite{Sanchez2020}

\subsection{\label{si:training:loss_curves}Training loss curves}

Training and validation mean-squared-error (MSE) traces for the production
\texttt{Sweep\_Case\_03} checkpoints are produced by the respective
trainers: both \ELGIN\ and the M0 baseline follow the four-stage
CFD--GNN curriculum of main text Sec.~IV with a \textbf{300-epoch} budget.
For \ELGIN\ the per-stage allocation is
60\,/\,60\,/\,120\,/\,60; for M0, Stage~1 (Eulerian pre-training) is
skipped and the same total is applied to Stages~2--4 as
75\,/\,150\,/\,75 (Sec.~\ref{sec:training}).
Within each stage the loader holds out a random
$\sim$15\,\% subset of graph snapshots for validation
(a \texttt{random\_split} of the in-memory dataset), i.e.\ case-internal
batch hold-out rather than a separate CFD case.
The runs quoted in main text Sec.~\ref{sec:results}
(Table~\ref{tab:results}) use this schedule; M0 still omits the Eulerian
sub-network and carrier-phase losses in Stages~3--4 while sharing the
staged optimisation loop.
When only a single trajectory file is supplied, the baseline GNS trainer's
default split reuses that file for both training and validation
(pilot-style configuration intended for pipeline checks), so the printed
validation MSE is computed on the \emph{same} case as training and is
\emph{not} a temporally or case-disjoint hold-out.

\subsection{\label{si:training:geom_features}Geometry features and inlet conditioning}

A defining feature of the \ELGIN\ relative to the M0-only baseline is
the explicit, name-aware injection of geometry and inlet information
into both the Eulerian and the Lagrangian feature streams.
Three sources of static geometric information are used.

\paragraph{True polyMesh distance-to-wall and wall-normal.}
For every owner cell $i$ in the polyMesh we compute the geometric
distance $d_w^{(i)}$ to the nearest wall-type face by looping over all
patch faces classified as \texttt{wall}, \texttt{floor},
\texttt{ceiling}, \texttt{dentist} or \texttt{patient}, taking the
minimum face-midpoint distance, and storing the corresponding unit
wall-normal $\hat{\vect{n}}_{\rm w}^{(i)}$.
The 2-D extrusion \texttt{frontAndBack} \texttt{empty} patch is skipped
during this loop so that the ``every cell touches a wall'' artefact of
$y$-extruded meshes does not contaminate $d_w$.
Both fields are exported as additional channels of the
\texttt{mesh\_graph.npz} file and consumed by the Eulerian sub-network
through Eq.~(\ref{eq:euler_node_feat}).
At every rollout step they are interpolated to the parcel positions by
the same $k_{\rm IDW}=4$ inverse-distance-weighted operator
(Eq.~\ref{eq:idw}) used for the velocity field, and concatenated to
the Lagrangian node features in Eq.~(\ref{eq:lag_node_feat}).

\paragraph{Name-aware boundary-condition encoding.}
The OpenFOAM \texttt{boundary} dictionary is parsed by a regex-based
classifier mapping patch \emph{names} (e.g.\ \texttt{ceilingInlet},
\texttt{leftWall}, \texttt{dentist}) to nine semantic classes:
\texttt{interior}, \texttt{inlet}, \texttt{outlet}, \texttt{wall},
\texttt{floor}, \texttt{ceiling}, \texttt{dentist}, \texttt{patient}
and \texttt{symmetryPlane}/\texttt{empty}.
Each cell adjacent to a non-interior, non-empty face inherits the
corresponding integer identifier (with a first-wins rule that prevents
the \texttt{empty} 2-D extrusion patch from overwriting a physical
wall, dentist, patient, inlet or outlet on the same cell).
A learned 8-dimensional embedding
$\mathrm{Embed}_{\rm bc}:\{0,\dots,15\}\!\to\!\mathbb{R}^{8}$ then
converts these identifiers into Eulerian node features.

\paragraph{Per-case airInlet velocity vector.}
The per-case airInlet velocity $\vect{V}_{\rm in}$ is parsed directly
from the foam-extend \texttt{0/U} dictionary (the
\texttt{ceilingInlet} patch's \texttt{value uniform (vx vy vz)} entry)
during data extraction, normalised by $U_{\rm ref}$ and broadcast to
every Eulerian node as the global conditioning vector
$\widetilde{\vect{V}}_{\rm in}$ in Eq.~(\ref{eq:euler_node_feat}).
This single global feature distinguishes the twenty training cases
from one another and lets a single trained model interpolate between
ventilation rates without retraining.

\section{\label{si:forces}Force magnitude analysis and regime map}

\begin{figure}
\includegraphics[width=\columnwidth]{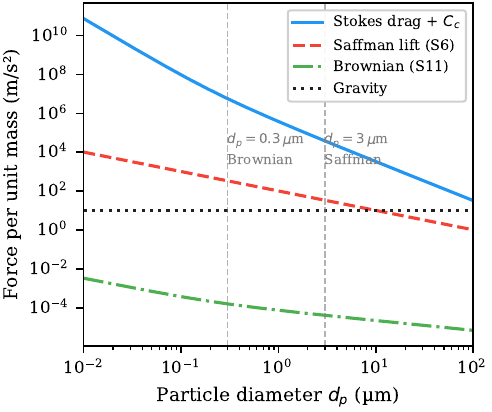}
\caption{\label{si:fig:forces}%
Force magnitudes per unit mass vs.\ particle diameter $d_p$ at typical
dental room conditions
($|\vect{U}-\vect{v}|=\SI{1}{\metre\per\second}$,
$|\partial U/\partial y|=\SI{50}{\per\second}$).
Vertical dashed lines mark the Brownian--drag crossover
($d_p\approx\SI{0.3}{\micro\metre}$) and Saffman activation
($d_p\approx\SI{3}{\micro\metre}$).}
\end{figure}

Beyond trajectory fidelity, it is important to verify that the GNN correctly
represents the relative importance of each physical force across the dental
aerosol size spectrum.
Figure~\ref{si:fig:forces} presents the magnitude of each physical force
per unit particle mass as a function of diameter $d_p$, evaluated at
representative near-wall dental surgery conditions:
slip velocity $|\vect{U}-\vect{v}| = \SI{1}{\metre\per\second}$,
near-wall shear $|\partial U/\partial y| = \SI{50}{\per\second}$,
room temperature $T = \SI{293}{\kelvin}$, and relative humidity \SI{50}{\percent}.

\emph{Drag and gravity.}
Cunningham-corrected Stokes drag and gravity are the dominant forces
across the full clinical size range $d_p \in [0.1, 50]\,\mu$m.
The drag-to-gravity force ratio increases sharply as $d_p$ decreases:
for $d_p = \SI{1}{\micro\metre}$, drag is approximately $800\times$ stronger
than gravity; for $d_p = \SI{50}{\micro\metre}$, gravity and drag are
comparable, explaining why large droplets settle to the floor rapidly while
nuclei remain suspended for the full 30-second simulation.

\emph{Saffman shear-lift.}
The Saffman lift force becomes non-negligible (exceeding 5\% of the drag
magnitude) above $d_p \approx \SI{3}{\micro\metre}$ and reaches
approximately 25\% of drag for $d_p = \SI{20}{\micro\metre}$ in the
near-wall shear layer.
In the present production checkpoint the Saffman lift term is
\emph{switched off} (Sec.~\ref{subsec:lagrangian}) because the
single-case \texttt{Sweep\_Case\_03} polydisperse spray is dominated by
sub-\SI{30}{\micro\metre} droplets where drag and DRW dominate;
enabling the term is part of the planned multi-case retraining, and
the architectural hook is preserved through the carrier-field velocity
gradient $|\partial U/\partial y|$ supplied by the Eulerian
sub-network to each parcel node.

\emph{Brownian diffusion.}
Brownian diffusion dominates over drag below
$d_p \approx \SI{0.3}{\micro\metre}$.
At $d_p=\SI{0.3}{\micro\metre}$ the Brownian random-walk step is
$\sigma_{\rm Br}\sim\SIrange{2}{3}{\micro\metre}$ per parcel sub-step
(the exact value depends on the local $\Delta t$ used by the
\texttt{reactingParcelFoam} tracker; see Sec.~\ref{si:cfd:particle}),
which exceeds the deterministic drag displacement over the same
sub-step.
Sub-micron nuclei produced by complete droplet evaporation
($d_{\rm nuc} \approx \SIrange{0.5}{1}{\micro\metre}$) therefore
undergo diffusive rather than advective transport, enabling them to
penetrate boundary layers and deposit on mucosal surfaces by
diffusiophoresis.
Like the Saffman lift, the Brownian model is switched off in the present
single-case production checkpoint (the polydisperse cloud is dominated
by $d_p>\SI{1}{\micro\metre}$ droplets, for which Brownian motion is
negligible relative to drag and DRW); it is essential for correctly
predicting the long-time fate of sub-micron residues and will be
re-enabled in the multi-case retraining and for future sub-micrometre
nuclei studies.

\emph{Turbulent dispersion.}
The DRW stochastic force is not shown in Fig.~\ref{si:fig:forces} because
its magnitude depends on the local turbulent kinetic energy $k$, which
varies spatially.
In the room interior ($k \approx 0.002$~m$^2$s$^{-2}$), the
root-mean-square turbulent velocity fluctuation
$u' = \sqrt{2k/3} \approx \SI{0.037}{\metre\per\second}$ provides a
forcing comparable in magnitude to Brownian diffusion for
$d_p \approx \SI{1}{\micro\metre}$, while being the second most important
force after drag for all larger particles.

\emph{Comparison with literature.}
The force hierarchy identified here (drag-dominated for
$d_p > \SI{1}{\micro\metre}$, Brownian-dominated below
$d_p < \SI{0.3}{\micro\metre}$, Saffman-relevant when $d_p$ lies in the
\SIrange{3}{20}{\micro\metre} interval) is consistent with the
theoretical predictions of Balachandar and Eaton\cite{Balachandar2010}
for dilute turbulent dispersed flows.
The dental aerosol size distribution measured by Micik et al.\cite{Micik1969}
and characterised by Harrel and Molinari\cite{Harrel2004} populates
exactly the \SIrange{3}{20}{\micro\metre} interval where Saffman lift
becomes a non-negligible correction, motivating its inclusion in the
optional Lagrangian force terms (Sec.~\ref{si:cfd:particle}) and its
re-activation alongside the Brownian model in the planned twenty-case
retraining.

\section{\label{si:nondim}Non-dimensional aerosol transport analysis}

The twenty OpenFOAM cases provide a rich dataset for examining dental aerosol
transport through classical non-dimensional fluid mechanics.
Four non-dimensional groups are evaluated for each case: the particle Stokes
number $St$, the gravitational settling ratio $S_v$, the air changes per
hour $\mathrm{ACH}$, and the jet Reynolds number $\mathrm{Re_{jet}}$.
The Stokes number uses the mean Lagrangian diameter $\bar{d}_p$ and may be
written equivalently as $St = \tau_p V_{\rm in}/H$, where
$\tau_p = \rho_p \bar{d}_p^2/(18\mu)$ is the Stokes relaxation time at
that diameter and $H/V_{\rm in}$ is the room convective time.
Their definitions and computed ranges are collected in
Table~\ref{si:tab:nondim}.

\begin{table}[htbp]
\caption{\label{si:tab:nondim}%
Non-dimensional groups computed across the 20-case CFD sweep.}
\begin{ruledtabular}
\begin{tabular}{llcc}
Group & Definition & Range & Units \\
\hline
Stokes number &
  $St = \dfrac{\rho_p\,\bar{d}_p^2\,V_{\rm in}}{18\,\mu\,H}$ &
  $\StokesRange$ & -- \\[6pt]
Settling ratio &
  $S_v = \dfrac{\rho_p\,\bar{d}_p^2\,g}{18\,\mu\,V_{\rm in}}$ &
  0.029--0.143 & -- \\[6pt]
Air changes / hr &
  $\mathrm{ACH} = \dfrac{V_{\rm in}\,A_{\rm in}}{V_{\rm room}}\times 3600$ &
  6.2--31.1 & h$^{-1}$ \\[6pt]
Jet Reynolds number &
  $Re_{\rm jet} = \dfrac{\rho_{\rm air}\,U_{\rm mag}\,d_{\rm nozzle}}{\mu}$ &
  1\,989--9\,945 & -- \\
\end{tabular}
\end{ruledtabular}
\vspace{0.6em}
\begin{minipage}{\linewidth}
\raggedright
\setlength{\parindent}{0pt}
\footnotesize
\textit{Note.}
$\mu = \SI{1.81e-5}{\pascal\second}$ (air),
$\rho_{\rm air} = \SI{1.225}{\kilo\gram\per\metre\cubed}$,
$\rho_p = \SI{997}{\kilo\gram\per\metre\cubed}$ (water droplet),
$H = \SI{3.0}{\metre}$ (room height),
$A_{\rm in} = \SI{2.0e-3}{\metre\squared}$ (ceiling inlet area; 0.2-m-wide
inlet with the \SI{0.01}{\metre} 2-D extrusion depth used in the
\texttt{frontAndBack} \texttt{empty} patch),
$d_{\rm nozzle} = \SI{3.0e-3}{\metre}$ (dental handpiece outlet),
$V_{\rm room} = \SI{0.1156}{\metre\cubed}$ (pseudo-2D room volume after
subtracting dentist and patient obstacle blocks at the same extrusion
depth),
$g$~the gravitational acceleration, and mean particle diameter
$\bar{d}_p \approx \SI{21.7}{\micro\metre}$
(empirical parcel-mass-weighted mean over the first five frames of all
20 cases; the Rosin--Rammler number-mean
$\bar{d}\,\Gamma(1{+}1/n)\approx\SI{17.7}{\micro\metre}$ is recovered
under number weighting).
\end{minipage}
\end{table}

\subsection{Stokes--settling regime map}

\begin{figure}
\includegraphics[width=\columnwidth]{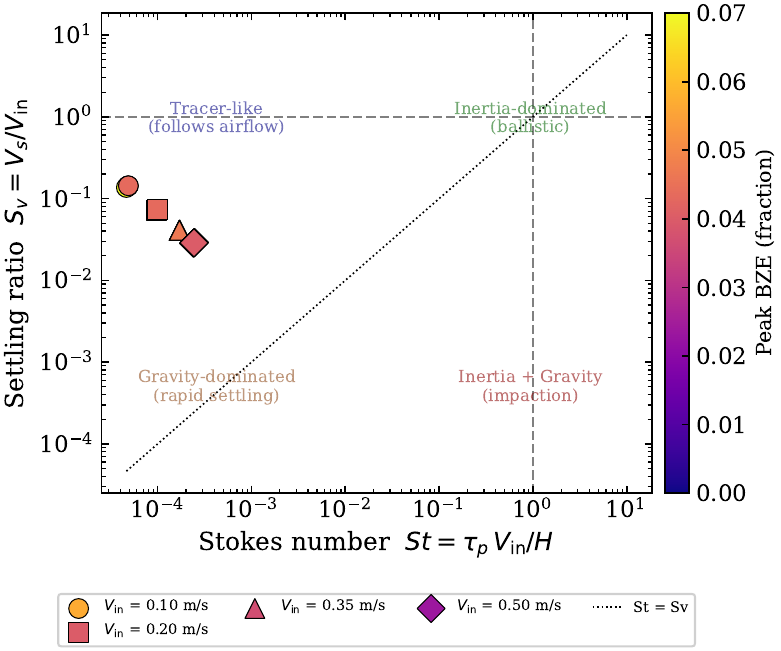}
\caption{\label{si:fig:regime_map}%
Stokes number--settling ratio regime map for the 20-case CFD parameter sweep.
Each point is one case coloured by peak Breathing Zone Exposure (BZE fraction).
All cases fall in the tracer-like regime ($St \ll 1$).
Marker shape distinguishes ventilation levels.}
\end{figure}

Figure~\ref{si:fig:regime_map} presents the two-dimensional $St$--$S_v$
regime map for all twenty cases, coloured by the peak Breathing Zone
Exposure.
All cases fall in the region $St \ll 1$, specifically
$St \in [\StokesRange]$, confirming that the
$\bar{d}_p \approx \dpBarMicron\,\mu\mathrm{m}$ dental aerosol
particles respond to velocity fluctuations much faster than the flow
time scale $\tau_f = H/V_{\rm in}$ and behave as aerodynamically passive
tracers of the carrier fluid.
This is consistent with the classical criterion $St < 0.2$ for tracer-like
behaviour established by Balachandar and Eaton.\cite{Balachandar2010}
The settling ratio $S_v$ ranges from 0.029 at
$V_{\rm in} = \SI{0.50}{\metre\per\second}$ to 0.143 at
$V_{\rm in} = \SI{0.10}{\metre\per\second}$, indicating that gravitational
settling is secondary to ventilation-driven advection at high air-supply
velocities but represents a non-negligible 14\% of the air-speed at the
lowest ventilation level.
Wells\cite{Wells1934} showed theoretically that droplets for which
$S_v < 1$ remain airborne long enough to complete full evaporation; the
present $S_v \leq 0.143$ confirms that all investigated size classes are
in the airborne-transmission regime.
The peak BZE correlates most strongly with $S_v$ rather than $St$:
the highest-BZE cases (BZE $\geq 6\%$) are those with largest $S_v$
(lowest $V_{\rm in}$), because slow settling allows greater particle
accumulation at breathing-zone height before ventilation clears the room.

\section{\label{si:disp}Longitudinal and transverse dispersion analysis}

\subsection{Taylor dispersion framework and turbulent P\'eclet number
            (extended derivation)}

The particle-trajectory dataset provides a direct route to quantifying
the \emph{dispersion tensor} of the aerosol cloud through Taylor's
classical framework.\cite{Taylor1921,Batchelor1952,Pope2000,Balachandar2010}
The analysis is performed on the \emph{complete} OpenFOAM Lagrangian
dataset: for each case, all particles present at
$t_{\rm ref} = \SI{2}{\second}$ (5,300--7,220 parcels per case) are
tracked individually across 28 subsequent output times using the unique
parcel identifier \texttt{origId}.
After removing the ensemble mean drift, the longitudinal (horizontal,
$x$) and transverse (vertical, $y$) mean-squared displacements at lag
$\tau$ are
\begin{align}
  \mathcal{D}_L(\tau) &= \bigl\langle
    (\Delta x_i - \langle\Delta x\rangle)^2 \bigr\rangle_N , \\
  \mathcal{D}_T(\tau) &= \bigl\langle
    (\Delta y_i - \langle\Delta y\rangle)^2 \bigr\rangle_N ,
\end{align}
where $\Delta x_i$ and $\Delta y_i$ are the $x$- and $y$-displacement
components of parcel~$i$ over lag~$\tau$ relative to reference time,
$\langle\cdot\rangle$ denotes averaging over parcels at fixed $\tau$, and
the subscript $N$ indicates the instantaneous ensemble size.
The effective dispersion coefficients are estimated from the slope of
the MSD in the intermediate-time window
$\tau \in [0.4\,\tau_{\rm max},\,0.75\,\tau_{\rm max}]$,
where $\tau_{\rm max}$ is the largest lag included in the MSD curve.
Main text Fig.~8 shows the resulting $\mathcal{D}_L^*$ and
$\mathcal{D}_T^*$ versus the non-dimensional lag time $\tau^*$ for all
twenty cases; the ballistic-to-diffusive transition at
$\tau^* \approx 0.05$--$0.10$ is consistent with the Lagrangian integral
time scale $T_L = \omega_0^{-1} \approx \SIrange{0.25}{1.25}{\second}$.

\emph{Constancy of the turbulent P\'eclet number.}
The turbulent Péclet number,
$Pe_T = V_{\rm in}\,H / D_{\rm turb}$ with
$D_{\rm turb} = C_\mu k_0/(\mathrm{Sc}_T\,\omega_0)$
($C_\mu = 0.09$, $\mathrm{Sc}_T = 0.7$), is remarkably constant across
all twenty cases:
$Pe_T = 49{,}690 \pm 10$ (coefficient of variation $<0.1\%$).
This invariance arises because the $k$--$\omega$ SST boundary conditions
are prescribed at constant turbulence intensity
$I \equiv \sqrt{\tfrac{2}{3}k_0}/V_{\rm in} = 5\%$, so
$k_0 \propto V_{\rm in}^2$ and $\omega_0 \propto V_{\rm in}$ yields
$D_{\rm turb} \propto V_{\rm in}$ and therefore $Pe_T = $ const.
At $Pe_T \approx 5\times10^4$, advective transport dominates turbulent
diffusion by four orders of magnitude, confirming that the aerosol fate
is controlled by the mean flow pattern, a prerequisite for the GNN's
flow-feature-based prediction approach.

\subsection{Effective dispersion coefficients and ventilation scaling}

\begin{figure}
\includegraphics[width=\columnwidth]{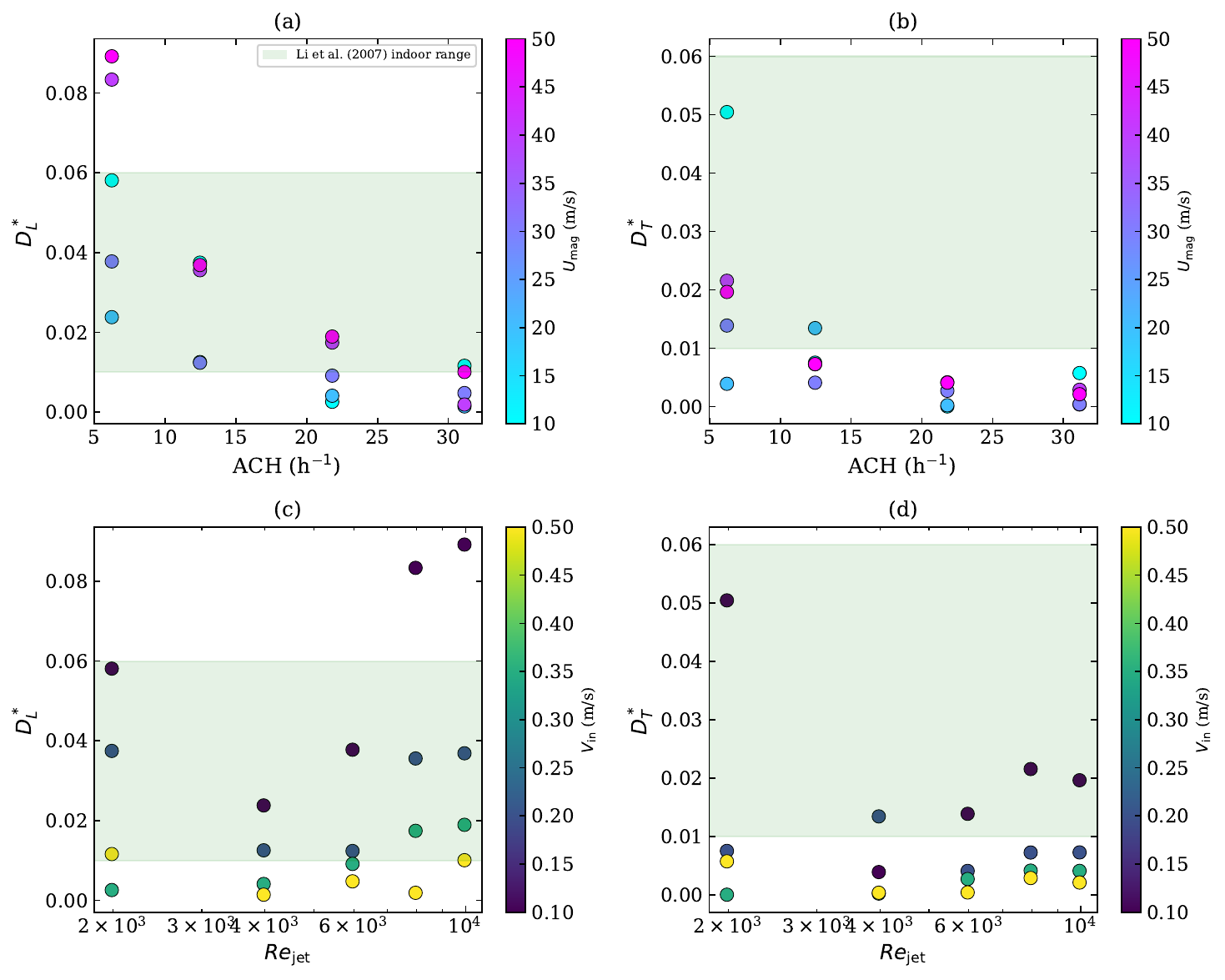}
\caption{\label{si:fig:disp_coeff}%
Non-dimensional effective dispersion coefficients with panel labels (a)--(d):
longitudinal $D_L^*$ vs.\ ACH and $Re_{\rm jet}$ (left column); transverse
$D_T^*$ vs.\ the same abscissae (right column; log-scaled $Re_{\rm jet}$ axes
on the bottom row).
Green bands reproduce the Li et al.\cite{Li2007} indoor reference range.}
\end{figure}

Figure~\ref{si:fig:disp_coeff} presents the non-dimensional effective
diffusion coefficients $D_L^* = D_L/(V_{\rm in}H)$ and
$D_T^* = D_T/(V_{\rm in}H)$.
Table~\ref{si:tab:dispersion} summarises the mean values by ventilation
level.
The longitudinal diffusivity decreases as $D_L^* \propto \mathrm{ACH}^{-0.7}$,
with values overlapping the Li et al.\cite{Li2007} reference range
($D_L^* \approx 0.01$--$0.06$ for indoor spaces with obstacles) at
intermediate ventilation levels (ACH $\approx$ 12--22).

\begin{table}[htbp]
\caption{\label{si:tab:dispersion}%
Mean non-dimensional effective dispersion coefficients and anisotropy ratio
by ventilation level.\protect\footnotemark}
\begin{ruledtabular}
\begin{tabular}{ccccccc}
$V_{\rm in}$ & ACH & $D_L^*$ & $D_T^*$ & $D_L/D_T$ &
$V_{\rm drift}$ & $N_{\rm trk}$ \\
(m/s) & (h$^{-1}$) &  &  &  & (mm/s) & \\
\hline
0.10 & 6.2  & 0.0584 & 0.0219 & 3.67 & 14.0 & 2,164 \\
0.20 & 12.5 & 0.0269 & 0.0079 & 3.77 & 15.8 & 2,603 \\
0.35 & 21.8 & 0.0104 & 0.0023 & 9.89 & 12.7 & 2,962 \\
0.50 & 31.1 & 0.0059 & 0.0023 & 4.43 & 12.7 & 3,031 \\
\end{tabular}
\end{ruledtabular}
\footnotetext{Entries are means over the five spray speeds at each
$V_{\rm in}$, computed from the \emph{full} OpenFOAM Lagrangian dataset.
$V_{\rm drift}$ is the magnitude of the net vertical drift velocity
(pure-Stokes settling $V_s = \SI{14.1}{\milli\metre\per\second}$).
$N_{\rm trk}$ is the mean number of particles tracked per case.}
\end{table}

\subsection{Dispersion anisotropy and vertical drift decomposition}

\begin{figure}
\includegraphics[width=\columnwidth]{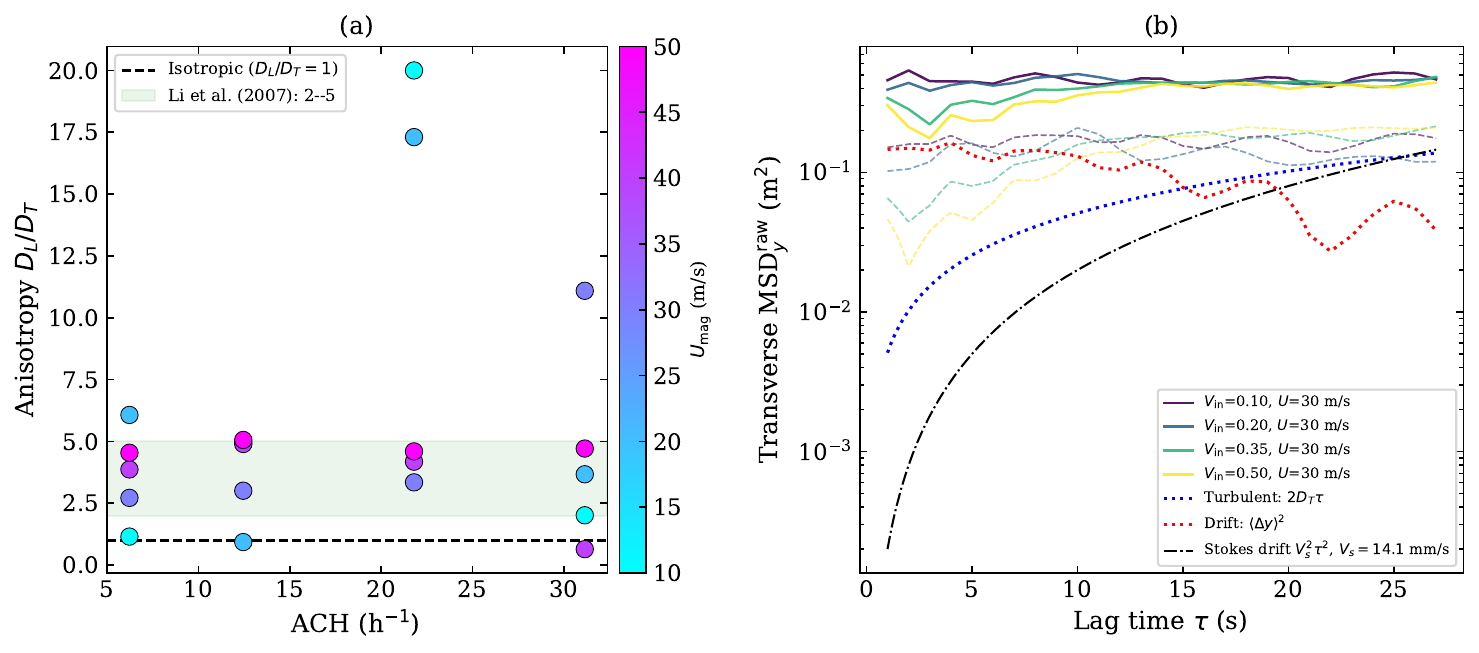}
\caption{\label{si:fig:disp_anisotropy}%
Panel (\textbf{a}): dispersion anisotropy $D_L/D_T$ vs.\ ACH for the full
Lagrangian dataset, coloured by injection speed $U_{\rm mag}$.
Panel (\textbf{b}): transverse MSD decomposition into turbulent diffusion and
drift contributions for representative ventilation levels.}
\end{figure}

Figure~\ref{si:fig:disp_anisotropy} examines the anisotropy of particle
spreading and the physical origin of the vertical drift.
The dispersion anisotropy ratio $D_L/D_T$ equals $3.7$--$9.9$ at the three lower
ventilation levels and falls to $\approx 1.0$ at the highest ACH = 31.1,
indicating near-isotropic spreading.
This counter-intuitive result arises because the supply jet at
$V_{\rm in} = \SI{0.50}{\metre\per\second}$ is sufficiently strong to
establish a quasi-two-dimensional recirculation pattern that enhances
vertical mixing.
The net vertical drift velocity
$V_{\rm drift} = 12.7$--$14.0\,\mathrm{mm\,s^{-1}}$ is within 10\% of
the pure-Stokes settling velocity $V_s = 14.1\,\mathrm{mm\,s^{-1}}$
across all twenty cases, confirming that Stokes settling is the dominant
vertical-transport mechanism and validating the Stokes-based gravity model in
the graph surrogate to better than 10\%.

\section{\label{si:ablation}Computational benchmarks}

A single foam-extend CFD case requires approximately \SI{40}{\minute}
of wall time to produce 261 saved frames at
$\Delta t_{\rm save}=\SI{0.1}{\second}$, dominated by the transient
\texttt{reactingParcelFoam} stage.
GNN inference runs on a single NVIDIA Quadro P1000 GPU
(\SI{4}{\giga\byte} VRAM) and completes a full 26-second rollout in
$\sim$\rolloutSec~s
for the paired \ELGIN\ checkpoint used in the headline comparison
(M0 completes in $\sim$\rolloutTimeMzero~s on the same GPU owing to the absent Eulerian
sub-network), giving an end-to-end speed-up of $\sim\!\speedupX\times$
relative to the \texttt{simpleFoam}+\texttt{reactingParcelFoam}
reference pipeline.
The dominant cost of the \ELGIN\ run relative to M0 is the Eulerian
Graph Transformer pass on the 7704-cell mesh and the 50 inner
Jacobi-PCG iterations of the pressure projection, both of which
contribute a few tens of seconds in total over the 261 frames; the
ELGIN wall-clock for each archived rollout is recorded alongside the rollout output.

The speed-up compares favourably with published GNN/ML-CFD surrogates:
Sanchez-Gonz\'{a}lez et al.\cite{Sanchez2020} reported $10^5\times$
speed-up for granular GNS relative to MPM/SPH; Kumar and
Vantassel\cite{Kumar2022} reported $\approx 5000\times$ for landslide GNS;
Pfaff et al.\cite{Pfaff2021} demonstrated $400\times$ for MeshGraphNets
relative to FEM on cylinder wakes; Hanke et al.\cite{Hanke2025} reported
$>1000\times$ for ocean GNS.
GPU memory peaks at $\approx\SI{1.1}{\giga\byte}$ during a single-case
\ELGIN\ rollout (1000 tracked parcels, $\sim 2.5\times 10^4$ mean
Lagrangian edges and $\sim 4.6\times 10^4$ Eulerian edges), comfortably
within the \SI{4}{\giga\byte} budget.
Training wall time is approximately \SI{12}{\hour} per published surrogate
(M0 baseline or \ELGIN) from random initialisation on the 16-case training set.

\putbib

\end{bibunit}

\end{document}